\documentclass{article}
\usepackage{PRIMEarxiv}
\usepackage[utf8]{inputenc} 
\usepackage[T1]{fontenc}    
\usepackage{hyperref}       
\usepackage{url}            
\usepackage{booktabs}       
\usepackage{amsfonts}       
\usepackage{nicefrac}       
\usepackage{microtype}      
\usepackage{lipsum}
\usepackage{fancyhdr}       
\usepackage{graphicx}       
\graphicspath{{media/}}     
\usepackage[dvipsnames]{xcolor}
\usepackage{algpseudocode}
\usepackage[inline]{enumitem}
\usepackage{tabularx}
\usepackage{placeins}
\usepackage{esvect}
\usepackage{amsmath,amsfonts,amssymb,amscd,bm}
\usepackage{subfigure}
\usepackage{algorithm}
\usepackage{makecell}
\usepackage{orcidlink}
\title{A Novel Hybrid Grey Wolf Differential Evolution Algorithm
}
\author{\\
Ioannis D. Bougas\textsuperscript{1, \orcidlink{0000-0002-7705-5865}}
Pavlos Doanis\textsuperscript{2, \orcidlink{0000-0002-5910-6710}},
Maria S. Papadopoulou\textsuperscript{1,3, \orcidlink{0000-0002-9651-2144}},
Achilles D. Boursianis \textsuperscript{1, \orcidlink{0000-0001-5614-9056}},\\
Sotirios P. Sotiroudis\textsuperscript{1, \orcidlink{0000-0003-3557-9211}},
Zaharias D. Zaharis\textsuperscript{4, \orcidlink{0000-0002-4548-282X}},
George Koudouridis\textsuperscript{1, \orcidlink{0000-0002-7372-5139}},
Panagiotis Sarigiannidis\textsuperscript{5, \orcidlink{0000-0001-6042-0355}},\\
Mohammad Abdul Matin\textsuperscript{6, \orcidlink{0000-0001-9312-4122}},
George Karagiannidis \textsuperscript{4,7, \orcidlink{0000-0001-8810-0345}},
Sotirios K. Goudos  \textsuperscript{1, 8, *, \orcidlink{0000-0001-5981-5683}}\\\\
{\small
\begin{tabular}{@{}p{0.5cm}p{14cm}@{}}
\textsuperscript{1} & ELEDIA@AUTH, ELEDIA Research Center, School of Physics, Aristotle University of Thessaloniki, 54124 Thessaloniki, Greece; impougas@physics.auth.gr (I.D.B.); bachi@physics.auth.gr (A.D.B.); ssoti@physics.auth.gr (S.S.); george.koudouridis@ieee.org (G.E.K.); sgoudo@physics.auth.gr (S.K.G.). \\
\textsuperscript{2} & Communication Systems Department, EURECOM, Sophia Antipolis, France; pavlos.doanis@eurecom.fr (P.D.). \\
\textsuperscript{3} & Department of Information and Electronic Engineering, International Hellenic University, Alexander Campus, GR-57400 Sindos, Macedonia, Greece; mspapa@ihu.gr (M.S.P.). \\
\textsuperscript{4} & Department of Electrical and Computer Engineering, Aristotle University of Thessaloniki, 54124, Greece; zaharis@auth.gr (Z.D.Z.); geokarag@auth.gr (G.K.). \\
\textsuperscript{5} & Department of Informatics and Telecommunications Engineering, University of Western Macedonia, 50131 Kozani, Greece; psarigiannidis@uowm.gr (P.S.). \\
\textsuperscript{6} & Department of Electrical and Computer Engineering, North South University, Dhaka, Bangladesh; mohammad.matin@northsouth.edu (M.A.M.). \\
\textsuperscript{7} & Cyber Security Systems and Applied AI Research Center, Lebanese American University (LAU), Lebanon. \\
\textsuperscript{8} & Department of Electronics and Communication Engineering, Bharath University, Chennai, Tamilnadu, India. \\
* & Correspondence: sgoudo@physics.auth.gr (S.K.G.). \\
\end{tabular}
}
}
\begin{document}
\maketitle

\begin{abstract}
{Grey wolf optimizer (GWO) is a nature-inspired stochastic meta-heuristic of the swarm intelligence field that mimics the hunting behavior of grey wolves. Differential evolution (DE) is a popular stochastic algorithm of the evolutionary computation field that is well suited for global optimization. In this part, we introduce a new algorithm based on the hybridization of GWO and two DE variants, namely the GWO-DE algorithm. We evaluate the new algorithm by applying various numerical benchmark functions. The numerical results of the comparative study are quite satisfactory in terms of performance and solution quality.}
\end{abstract}


\section{Introduction}\label{gwo1}
EAs are evolutionary algorithms, which are stochastic optimizers. Fundamentally, EAs emulate the evolution and behavior of biological entities, drawing inspiration primarily from Darwin's theory of evolution and the mechanism of natural selection. Further benefits of EAs include their frequent compatibility with deterministic methods and their ease of parallelization across multiple cores or CPUs as a result of their evolving population.

Swarm Intelligence (SI) algorithms fall under the category of Evolutionary Algorithms (EAs) in particular. The collective behavior of self-organized and decentralized swarming is referred to as SI. Particle Swarm Optimization (PSO) \cite{RN57}, Ant Colony Optimization (ACO) \cite{RN1220, RN1219, RN346}, and Artificial Bee Colony (ABC) \cite{RN951} are all instances of SI algorithms. PSO emulates the schooling and flocking behaviors of fish and birds, respectively \cite{RN57}. According to \cite{RN103}, the two PSO algorithms that are most frequently implemented are Inertia Weight PSO (IWPSO) and Constriction Factor PSO (CFPSO). Real-world problems are amenable to PSO due to its computational efficacy and simplicity of implementation. ACO emulates the conduct of authentic ants, whereas ABC simulates the foraging behavior of honey bees. Furthermore, the Swarm Intelligence domain encompasses a number of nascent algorithms, including the Grey Wolf Optimizer (GWO) \cite{MIRJALILI2014}, the Whale Optimization Algorithm (WOA) \cite{WOA}, and the Salp Swarm Algorithm (SSA) \cite{Mirjalili2017}. The Monarch butterfly optimization (MBO) \cite{MBO}, the moth search algorithm (MSA) \cite{MSA}, the Firefly Algorithm (FA) \cite{Yang2009}, and the Elephant Herding Optimization (EHO) \cite{EHO} are additional emerging algorithms inspired by nature.

The grey wolf optimizer (GWO) \cite{GWO} is a meta-heuristic optimization algorithm that draws inspiration from the fundamental behavior of grey wolves.
Differential evolution (DE) \cite{RN91} is an algorithm for global optimization that is based on a population and operates in a stochastic manner. It has been applied to various engineering problems in real-world scenarios, utilizing different versions of the DE algorithm. While lacking direct inspiration from any particular natural phenomenon, DE is constructed using a mathematical model of evolution as its foundation.
In the literature, several different DE algorithms or strategies depend on the way mutant vectors are generated. A popular self-adaptive DE algorithm has been reported in \cite{jDE2006}. More details on the application of EAs to antennas and wireless communications can be found in \cite{goudos2021emerging}.

In this work, we propose a new hybrid GWO-DE algorithm that combines the GWO with two variants of DE, the \emph{ DE / best / 1 / bin} and the self-adaptive jDE algorithm. The new hybrid algorithm changes the current algorithm used by evaluating the performance of the last iterations. That way the GWO-DE tries to avoid stagnation.  We evaluated the new algorithm applying common numerical benchmark functions, as well as the CEC 2017 test suite \cite{CEC2017}.

The structure of this paper is as follows: Firstly, we provide a description of the algorithms in Section \ref{gwoalgorithm}. In Section \ref{gwores}, we present the numerical results obtained by applying eight different EAs to numerical benchmark problems. Finally, Section \ref{gwoConcl} summarizes the concluding remarks of our work.
\section{Algorithm Description}\label{gwoalgorithm}
\subsection{GWO}
The GWO algorithm is designed on the basis of mathematical representations of the fundamental mechanisms (hunting and hierarchy) of grey wolves in their natural habitat. Its primary attribute is its ability to retain information on the search space throughout the iterative process. No control parameters are mandatory in GWO. The algorithm partitions the population into four categories, with alpha ($\alpha$), beta ($\beta$), and delta ($\delta$) representing the top three solutions/grey wolves.

Omega-classified ($\omega$) solutions comprise the entirety of the category. Consequently, group foraging (optimization process) is influenced by the aforementioned population categories ($alpha$, $beta$, $delta$), similar to social behavior in a pack of wolves. The mathematical formulation illustrating the process of prey encirclement during hunting is as follows:
\begin{equation}\label{eq01}
\vv{V}=|\vv{C}^{2}\cdot\vv{P}_{G}-\vv{W}_{G}|
\end{equation}
\begin{equation}\label{eq02}
\vv{W}_{G+1}=\vv{P}_{G}-\vv{C}^{1}\cdot\vv{V}
\end{equation}
\noindent where $\vv{C}^{1}$ and $\vv{C}^{2}$ are coefficient vectors, $\vv{P}$ is the position vector of the prey, $\vv{W}$ corresponds to the position vector of the grey wolf, and $G$ indicates the current generation.
Vectors $\vv{C}^{1}$ and $\vv{C}^{2}$ are given by \eqref{eq03} and \eqref{eq04}:
\begin{equation}\label{eq03}
\vv{C}^{1}=2\vv{u}\cdot\vv{v}_{1}-\vv{u}
\end{equation}
\begin{equation}\label{eq04}
\vv{C}^{2}=2\cdot\vv{v}_{2}
\end{equation}
\noindent where $\vv{u} \in [2,0]$ and $\vv{v}_{1}, \vv{v}_{2} \in [0,1]$ (random vectors).
Compared to the social behavior of a flock of grey wolfs, the GWO algorithm's hunting process can be characterized by the subsequent set of equations.
\begin{eqnarray}\label{eq05}
\begin{array}{l}
\vv{V}_{\alpha}=|\vv{C}_{1}^{2}\cdot\vv{W}_{\alpha}-\vv{W}|\\
\vv{V}_{\beta}=|\vv{C}_{2}^{2}\cdot\vv{W}_{\beta}-\vv{W}|\\
\vv{V}_{\delta}=|\vv{C}_{3}^{2}\cdot\vv{W}_{\delta}-\vv{W}|
\end{array}
\end{eqnarray}
\begin{eqnarray}\label{eq06}
\begin{array}{l}
\vv{W}_{1}=\vv{W}_{\alpha}-\vv{C}_{1}^{1}\cdot\big(\vv{V}_{\alpha}\big)\\
\vv{W}_{2}=\vv{W}_{\beta}-\vv{C}_{2}^{1}\cdot\big(\vv{V}_{\beta}\big)\\
\vv{W}_{3}=\vv{W}_{\delta}-\vv{C}_{3}^{1}\cdot\big(\vv{V}_{\delta}\big)
\end{array}
\end{eqnarray}
\begin{equation}\label{eq07}
\vv{W}_{G+1}=\frac{\vv{W}_{1}+\vv{W}_{2}+\vv{W}_{3}}{3}
\end{equation}
The pseudo-code of GWO algorithm is outlined in \ref{gwo}.
\begin{algorithm}[h]
\caption{GWO algorithm}
\label{gwo}
\begin{algorithmic} [1]
\State Initialize the population of grey wolves $\vv{W}_{i}(i=1,2,\cdots,n)$
\State Initialize $\vv{u}$, $\vv{C}^{1}$, and $\vv{C}^{2}$
\State Calculate the category position vectors: $\vv{W}_{\alpha}$, $\vv{W}_{\beta}$, and $\vv{W}_{\delta}$ \\
\While {($t<$ max no. of iterations)}
\For {each member of the pack}
\State Update the position vector of the current member by \eqref{eq07}
\EndFor
\State Update $\vv{u}$, $\vv{C}^{1}$, and $\vv{C}^{2}$
\State Calculate the position vectors of all members
\State Update $\vv{W}_{\alpha}$, $\vv{W}_{\beta}$, and $\vv{W}_{\delta}$
\State Set: $t=t+1$
\EndWhile
\end{algorithmic}
\end{algorithm}
\FloatBarrier

\subsection{DE}
Differential evolution (DE) is a widely utilized evolutionary algorithm\cite{Sto1995} \cite{Storn1997}. As a mathematical construction, DE does not represent any natural phenomenon.

DE algorithms evolve the population with the assistance of three operators per generation. Mutation, crossover, and selection are examples. The literature contains references to a number of distinct DE variants \cite{Storn1997,Storn20081}. The differentiation among these variations resides in the mathematical expression used to express the DE operators.
The selection of the most appropriate DE strategy or variant to implement is contingent upon the characteristics of the particular optimization problem \cite{Mezura-Montes2006}. By employing the notation \emph{DE/x/y/z}, we determine the vectors to be mutated based on the following criteria: $x$ represents the vector to be mutated (\emph{"best"} denotes the optimal vector and \emph{"rand"} represents a random vector); $y$ represents the quantity of vectors of type $x$ that are mutated; and $z$ signifies the crossover type (typically \emph{"bin"} for binary or \emph{"exp"} for exponential).
The \emph{DE/rand/1/bin} and \emph{DE/best/1/bin} strategies are two of the most widely recognized D.

Then, we can formulate the $m$-th mutant vector ${\bar V_{t ,m}}$ for each target vector ${\bar X_{t,m}}$ in the $t$ iteration as :
\begin{eqnarray}
\begin{array}{l}
\emph{DE/rand/1/bin}\\
{{\bar V}_{t,m}} = {{\bar X}_{t,{r_1}}} + F \times ({{\bar X}_{t,{r_2}}} - {{\bar X}_{t,{r_3}}}),{r_1} \ne {r_2} \ne {r_3}\\
\emph{DE/best/1/bin}\\
{{\bar V}_{t,m}} = {{\bar X}_{t,best}} + F\times ({{\bar X}_{t,{r_1}}} - {{\bar X}_{t,{r_2}}}),{r_1} \ne {r_2}
\end{array}
\end{eqnarray}
\noindent where ${r_1},{r_2},$ and ${r_3}$ are randomly selected population members, and $F$ is the mutation control parameter or mutation scale factor.
Then, we use the crossover operator to generate a trial vector ${\bar U_{t,m}}$ written as:
\begin{eqnarray}
\label{DECR}
{U_{t,km}} = \left\{ \begin{array}{l}
{V_{t,km}}, if ran{d_{k[0,1)}} \le CR\,\,{ or\,\, k = rn(m)}\\
{X_{t,km}},{    \,if\, }ran{d_{k[0,1)}}{ > }CR\,\,{and\,\, k} \ne {rn(m)}
\end{array} \right.{}
\end{eqnarray}
\noindent where $k = 1,2,......,D{,   }ran{d_{k[0,1)}}$ is a uniformly distributed random number in the interval [0,1), ${rn(m)}$ a randomly chosen index from $(1,2,......,D)$,  and $CR$ another DE control parameter, the crossover constant from the interval $[0,1]$.
Moreover, the DE algorithms use  a greedy selection operator, which for minimization problems is expressed as:
\begin{eqnarray}
\label{greedy}
{\bar X_{t + 1,m}} = \left\{ \begin{array}{l}
{{\bar U}_{t,m}}{,    if \,\,}f({{\bar U}_{t,m}}) < f({{\bar X}_{t,m}})\\
{{\bar X}_{t,m}},{      \,\,otherwise}
\end{array} \right.{}
\end{eqnarray}
\noindent where $f({\bar U_{t,m}})$, $f({\bar X_{t,m}})$ are the objective function values of the trial and the old vector respectively.
Therefore, the new trial vector ${\bar U_{t,m}}$ replaces the old vector ${\bar X_{t,m}}$  if it is better than the previous one. Otherwise, the old vector remains in the population.
The configuration of the crossover and mutation control parameters $F$ and $CR$ is necessary for DE. Therefore, the fact that only two control parameters are necessary is advantageous for DE.
However, refining these control parameters might necessitate an arduous process of experimentation and trial. The setting of the differential evolution control parameters $F$ and $CR$ from the intervals $[0.5,1]$ and $[0.8,1]$, respectively, is recommended in \cite{Storn1997}.
In light of this, numerous adaptive and self-adaptive DE variants have been suggested in the academic literature.

\subsubsection{Self-adaptive DE}

In \cite{Bre2006} proposes a DE strategy (dubbed jDE by the authors) in which the parameters $F$ and $CR$ self-adapt.
In \cite{Bre2006}, the authors introduce a self-adaptive approach referred to as jDE. The jDE variant employs the identical mutation operator that is utilized in the \emph{DE/rand/1/bin} scheme.
The jDE algorithm operates on the premise that each vector should contain two additional unknowns. These are the individual's $F$ and $CR$ values. These are the control parameters that jDE develops. The jDE strategy generates novel vectors by adjusting the values of the control parameters. These vectors are believed to have a higher probability of survival and proliferation in the subsequent generation. Therefore, the values of the improved control parameters are retained by subsequent generations of new superior vectors.
The control parameters for each vector of the population are self-adjusted in each generation using the following formulas:
\begin{eqnarray}
\label{eq:jDE}
\begin{array}{l}
{F_{t + 1,m}} =\left\{ \begin{array}{l}
{F_l} + rn{d_{1[0,1]}} \times {F_u}\,\,{if}\,rn{d_{2[0,1]}} < {prob_1}\\
{F_{t,m}},\,\,{otherwise}
\end{array} \right.{}\\
C{R_{t + 1,m}} =\left\{ \begin{array}{l}
rn{d_{3[0,1]}}\,\,{if}\,rn{d_{4[0,1]}} < {prob_2}\\
C{R_{t,m}},\,{otherwise}
\end{array} \right.
\end{array}
\end{eqnarray}
\noindent where $rn{d_{k[0,1]}},{}k = 1,2,3,4$ represent  uniform random numbers $ \in \left[ {0,1} \right]$, ${F_l},{F_u}$ are the lower and the upper limits of $F$ set to 0.1 and 0.9, respectively, and ${prob_1}$ and ${prob_2}$ the probabilities of adjusting
the control parameters.
In the original jDE paper, the authors set both probabilities at $0.1$ after several trials. Additionally, the conclusion from the performance of the jDE strategy is that it is better or at least comparable to the classical DE \emph{DE/rand/1/bin} strategy \cite{Bre2006}. Furthermore, the use of jDE does not increase the time complexity. The interested reader may seek more details on the jDE strategy in \cite{Bre2006}.
\subsection{GWO-DE}
In this paper, we introduce a new hybrid GWO-DE algorithm that combines GWO with two DE variants, the \emph{DE/best/1/bin} strategy, and the self-adaptive jDE algorithm.

Similar to \cite{Hybrid} GWO-DE uses the concept of
the average number of unsuccessful updates to control the execution of GWO, jDE or \emph{DE/best/1/bin}.
This new hybrid algorithm uses three new control flags named $Q_{1},Q_{2},Q_{3}$. These flags obtain predefined values, and their main use is to avoid stagnation. Their role is to switch between the base algorithms (i.e., GWO and the two DE variants) in case the current result does not improve after each generation. The algorithm initiates using GWO; a $Q$ indicator counts the number of unsuccessful generations that do not improve the value of the objective function. If $Q$ becomes larger than $Q_{1}$, then the algorithm switches to \emph{DE/best/1/bin}, and the $Q$ is set back to zero. Again, the algorithm counts the number of unsuccessful generations. If $Q$ becomes larger than $Q_{2}$, then the algorithm switches to the jDE algorithm and the $Q$ is set back to zero. Finally, if jDE is unsuccessful (by means of failing to improve the objective function value), the hybrid algorithm switches back to GWO after $Q_{3}$ unsuccessful generations. The pseudo-code of the proposed hybrid algorithm is outlined in \ref{algorithm:GWO-DE}.
\begin{algorithm} [h]
\caption{Pseudo-code of GWO-DE algorithm.}
\label{algorithm:GWO-DE}
\begin{algorithmic}[1]
\State{Define the population number $NP$ of the optimization problem}
\State{Define the number of iterations $MaxIt$ of the optimization process}
\State{Define the number of decision variables $MaxVar$ of the optimization problem}
\State{Define the fitness function $F(x)$ and the best fitness function $F_{Best}(x)$ of the optimization problem}
\State{Define the control parameters $Q_{1}$, $Q_{2}$, $Q_{3}$ of the optimization process}
\State Set $Q=0$
\For{$i=1$ to $MaxIt$}
\For{$j=1$ to $NP$}
\For{$k=1$ to $MaxVar$}
\If{$F_{Best}^G(x)>=F_{Best}^{G-1}(x)>$}
\State{$Q=Q+1$}
\EndIf
\If{$Q>Q_{1}$}
\State{Set $flag=2$, $Q=0$}
\EndIf
\If{$Q>Q_{2}$}
\State{Set $flag=3$, $Q=0$}
\EndIf
\If{$Q>Q_{3}$}
\State{Set $flag=1$, $Q=0$}
\EndIf
\State{------GWO algorithm execution------}
\If{$flag=1$}
\State{Execute the GWO algorithm}
\EndIf
\State{------DE/best/1/bin strategy execution------}
\If{$flag=2$}
\State{Execute the \emph{DE/best/1/bin} strategy of the DE algorithm}
\EndIf
\State{------jDE algorithm execution------}
\If{$flag=3$}
\State{Execute the jDE algorithm}
\EndIf
\EndFor
\EndFor
\EndFor
\end{algorithmic}
\end{algorithm}
\FloatBarrier
The time complexity of the proposed GWO-DE algorithm in each iteration is $\mathcal{O}(NPD + NPF)$, where $NP$ is the population size, $F$ is the time complexity of the objective function, and $D$ is the dimensions of the problem. The "if...else" statements in GWO-DE do not affect the overall time complexity. This results in the same time complexity, as in the one produced by the combination of the DE and the GWO.
\section{Numerical results.}\label{gwores}

\subsection{Benchmark functions}

We apply the new GWO-DE algorithm to eleven well-known (Schaffer $f_{6}$, Sphere, Rosenbrock, Ackley, Gen. Griewank, Weierstrass, Gen. Rastrigrin, Non-cont. Rastrigrin, Gen. Penalized $f_{12}$, and Gen. Penalized $f_{13}$) numerical benchmark functions, and we compare its performance against seven legacy algorithms, the original GWO \cite{GWO}, the \emph{DE/best/1/bin} \cite{DE}, the jDE \cite{jDE2006}, the Jaya algorithm \cite{Rao2016}, the Teaching-Learning-Based Optimization (TLBO) \cite{Rao2011}, the Particle Swarm Optimization (PSO) \cite{RN57}, and the Artificial Bee Colony (ABC) \cite{RN951}. In all cases, the number of decision variables is set to $30$.

All algorithms are applied for $50$ independent trials; the population size is set to $200$ and the maximum number of iterations is $1000$. Therefore, the total number of evaluations of the objective function is $200,000$. Table \ref{tab1CTF} reports the comparative results of the algorithms. GWO-DE achieves the best value in five out of the eleven cases, which is the best performance of all algorithms.

Table \ref{tab2CTF} lists the numerical results obtained by applying the Friedman non-parametric test. From Table \ref{tab2CTF}, we can easily derive that the proposed GWO-DE emerges as the best algorithm in terms of average ranking.

\begin{figure}[H]%
\centering
\subfigure[\label{fig:firstA0}]{\includegraphics[height=1.5in]{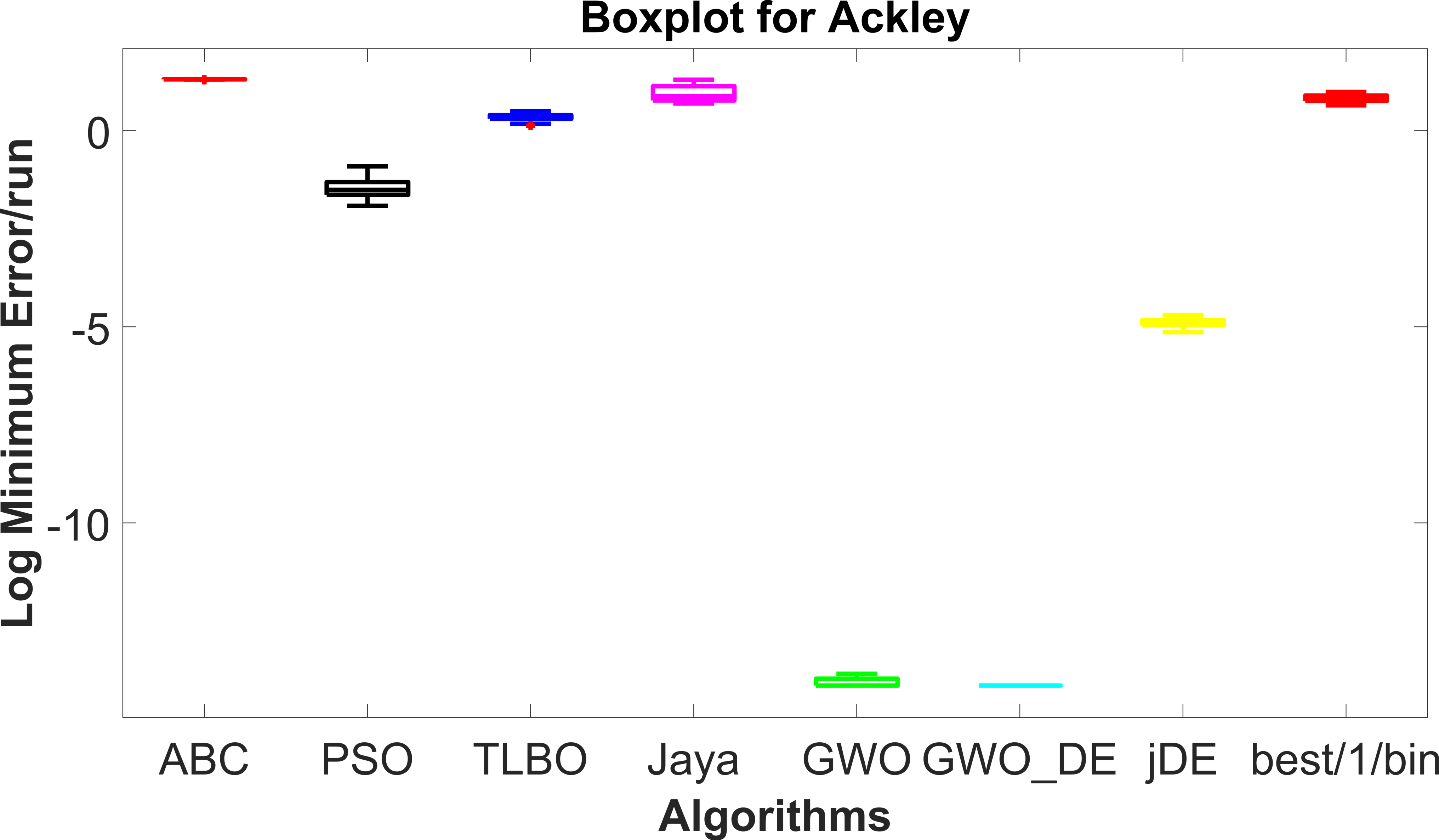}}%
\hfill
\subfigure[\label{fig:firstB0}]{\includegraphics[height=1.5in]{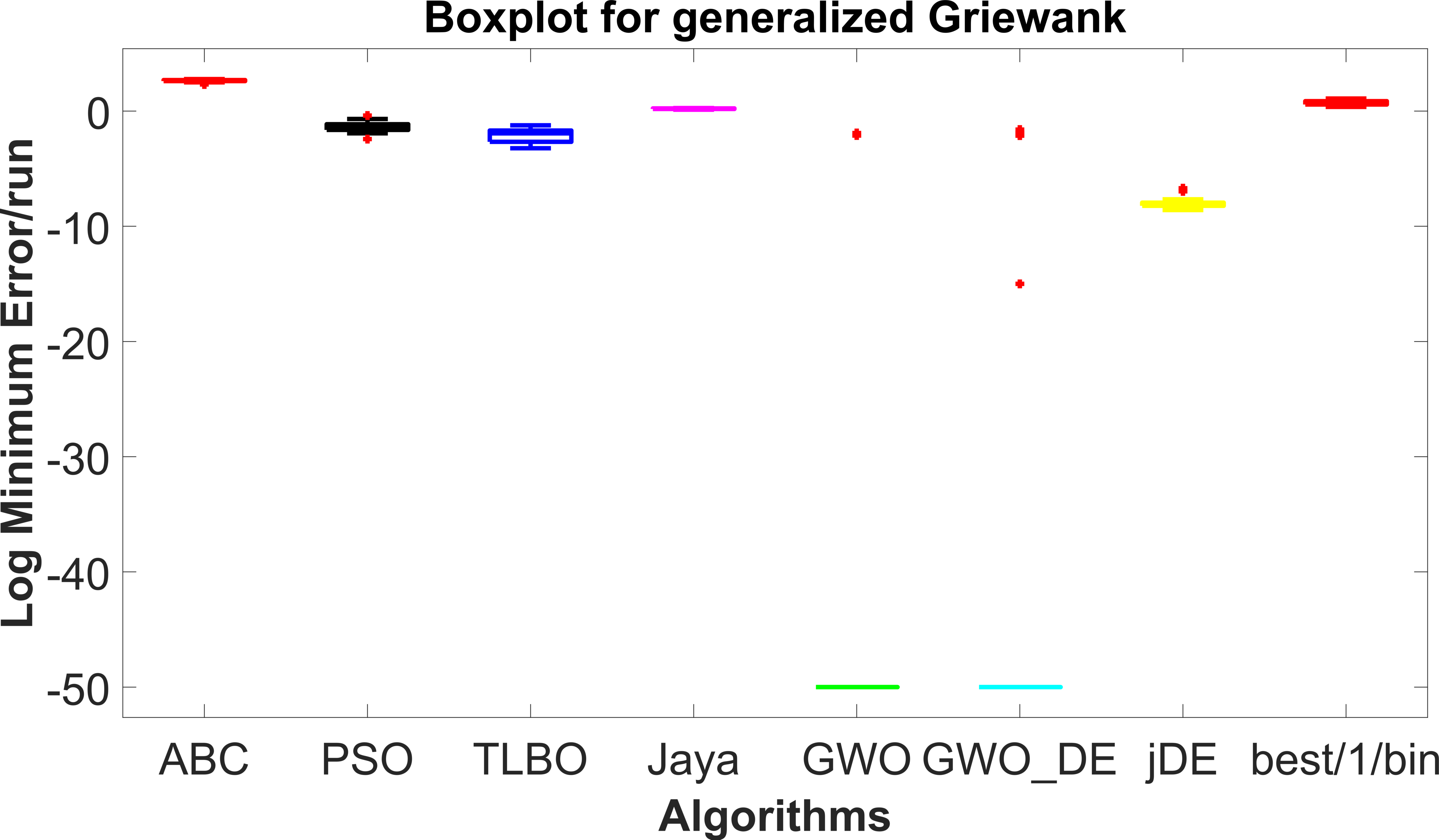}}%
\hfill
\subfigure[\label{fig:firstC0}]{\includegraphics[height=1.5in]{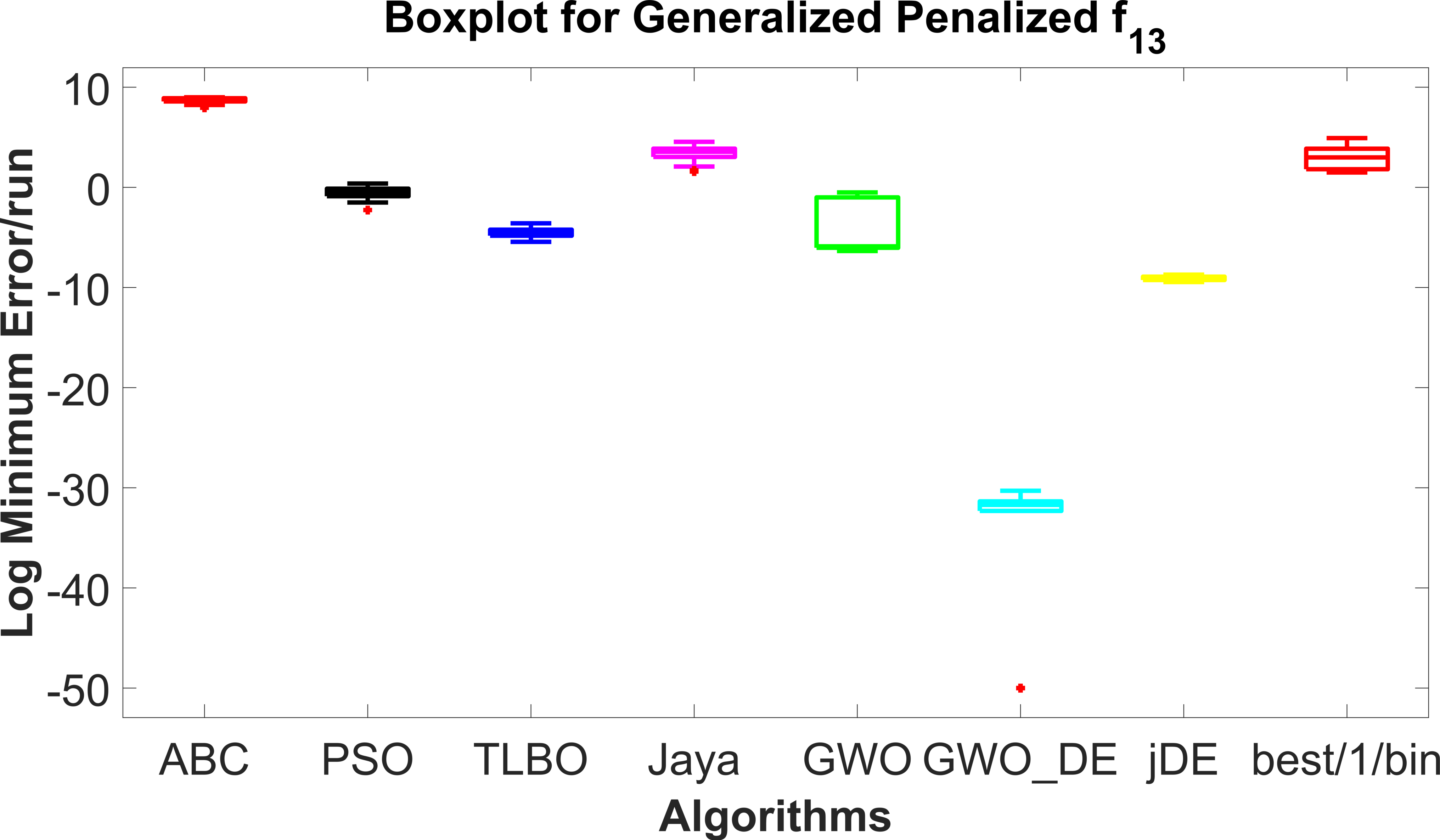}}%
\hfill
\subfigure[\label{fig:firstD0}]{\includegraphics[height=1.5in]{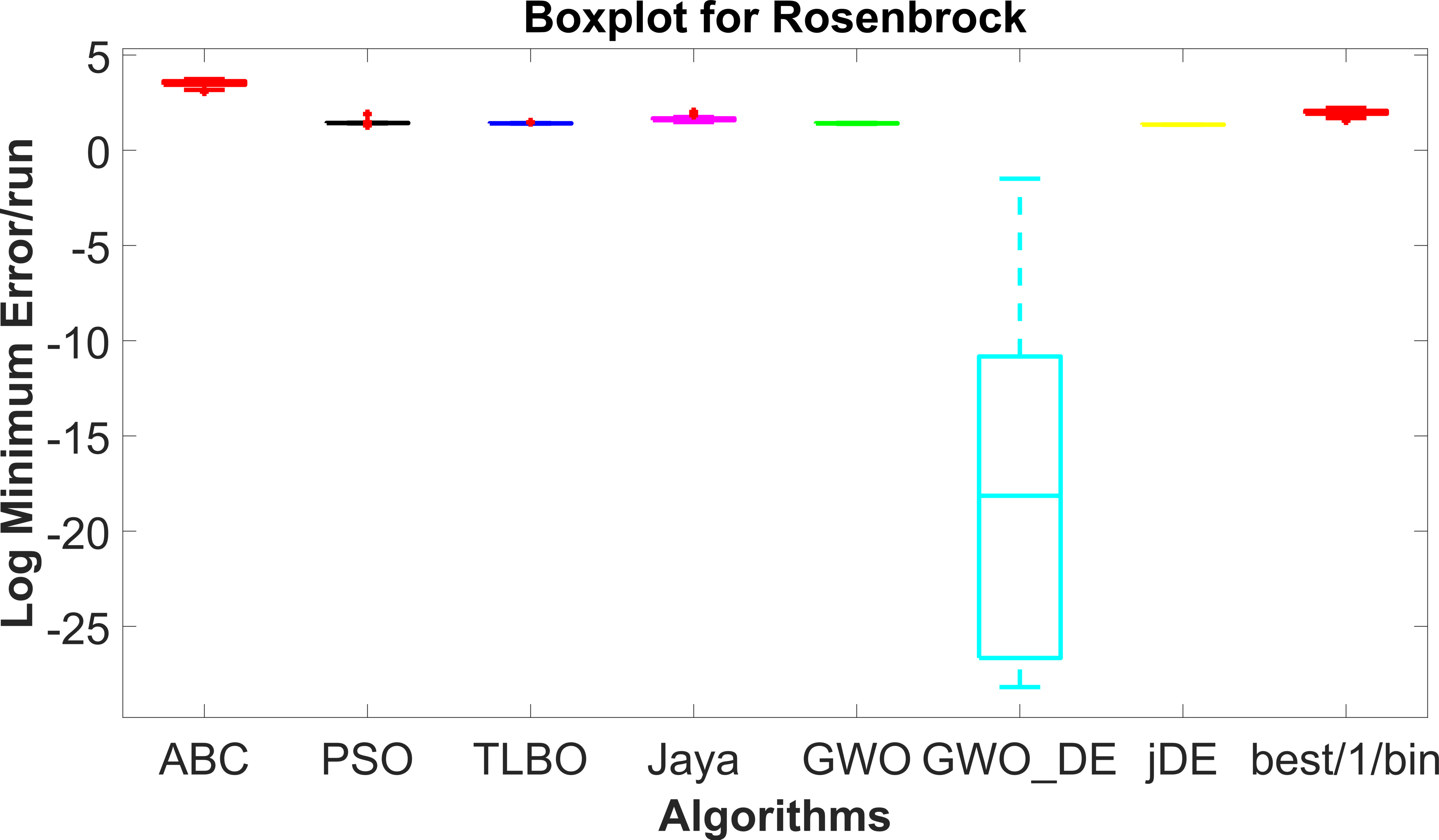}}%
\caption{Box plots of log error distribution for the common  benchmark functions, $D=30$, a) Ackley, b) generalized Griewank, c) Generalized Penalized $f_{13}$, d) Rosenbrock.}
\label{fig:boxCTF30}%
\end{figure}

\begin{table}[h]
\caption{Comparative results of common benchmark functions. The bold font indicates the smaller value.}
\label{tab1CTF}
\newcolumntype{C}{>{\centering\arraybackslash}X}
\begin{tabularx}{\textwidth}{|C|C|C|C|C|C|C|C|C|}
\hline\hline
\textbf{Test   function} &\textbf{ABC} &\textbf{PSO} &\textbf{TLBO} &\textbf{Jaya} &\textbf{GWO} &\textbf{GWO-DE} &\textbf{jDE} &\textbf{{DE/
best
/1/
bin}} \\
\hline
Schaffer
\_f6        & 1.84
E-01     & \textbf{0.00
E+00} & 5.95
E-12      & 1.12
E-04      & 1.94
E-04          & 3.89
E-04          & \textbf{0.00
E+00} & \textbf{0.00
E+00} \\ \hline
Sphere   & 5.22
E+04     & 1.38
E-06          & 9.90
E-04      & 6.74
E+01      & 7.86
E-99          & \textbf{1.32
E-99} & 2.04
E-09          & 6.24
E+02 \\ \hline
Rosen-
brock      & 3.26
E+03     & 2.77
E+01          & 2.57
E+01      & 4.70
E+01      & 2.59
E+01          & \textbf{7.72
E-05} & 2.20
E+01          & 9.55
E+01    \\ \hline
Ackley                       & 2.03
E+01     & 3.49
E-04          & 2.25
E+00      & 7.64
E+00      & 7.32
E-15          & \textbf{7.03
E-15} & 1.34
E-05          & 7.11
E+00  \\ \hline
Genera-
lized
Grie-
wank        & 4.51
E+02     & 1.49
E-02          & 1.05
E-02      & 1.60
E+00      & 1.27
E-03          & 1.28
E-03          & \textbf{1.49
E-08} & 5.64
E+00 \\ \hline
Weier-
strass                  & 3.55
E+01     & 1.24
E-02          & 3.71
E-01      & 1.34
E+01      & 6.23
E-01          & 4.67
E-01          & \textbf{1.50
E-03} & 1.14
E+01  \\ \hline
Genera-
lized Rastri-
grin      & 3.77
E+02     & 1.91
E+01          & 3.07
E+00      & 2.53
E+02      & \textbf{0.00
E+00} & 2.63
E+00          & 4.73
E+01          & 7.64
E+01  \\ \hline
Noncon-
tinuous Rastri-
grin   & 3.42
E+02     & 2.30
E+01          & 5.52
E+00      & 2.21
E+02      & \textbf{2.80
E-01} & 2.78
E+00          & 3.03
E+01          & 7.30
E+01 \\ \hline
Genera-
lized Pena-
lized f12  & 2.17
E+08     & 6.83
E-03          & 4.35
E-02      & 3.49
E+01      & 8.61
E-03          & \textbf{1.07
E-31} & 2.00
E-10          & 5.67
E+01  \\ \hline
Genera-
lized
Pena-
lized f13 & 5.35
E+08     & 1.61
E-05          & 2.65
E-04      & 5.40
E+03      & 6.88
E-02          & \textbf{4.02
E-32} & 8.25
E-10          & 6.68
E+03 \\
\hline\hline
\end{tabularx}
\vspace{5pt}
\end{table}
\FloatBarrier
\begin{table}[h]
\caption{Average Rankings achieved by Friedman test for common benchmark functions.\label{tab2CTF}}
\newcolumntype{C}{>{\centering\arraybackslash}X}
\begin{tabularx}{\textwidth}{|C|C|C|C|}
\hline\hline
\textbf{Algorithm}	& \textbf{Average Ranking}	& \textbf{Normalized} & \textbf{Ranks}\\
\hline
ABC & 8.00     & 3.50   & 8 \\\hline
PSO & 3.60   & 1.60   & 4  \\\hline
TLBO & 3.90   & 1.70   & 5  \\\hline
Jaya  & 6.30   & 2.70   & 7 \\\hline
GWO  & 3.20   & 1.40   & 3  \\\hline
GWO-DE & \textbf{2.30} & \textbf{1.00}   & \textbf{1} \\	\hline
jDE & 2.60   & 1.10   & 2  \\\hline
\emph{DE/best/1/bin} & 6.10   & 2.70   & 6 \\
\hline\hline
\end{tabularx}
\vspace{5pt}
\end{table}
\FloatBarrier
Figure \ref{fig:boxCTF30} illustrates the box plots of four different function error distributions. For visualization purposes, the error was converted to $log_{10}$. We notice that GWO-DE obtains the smaller distribution, except for the Rosenbrock function. However, in this case, GWO-DE obtained the best error values.

\begin{figure}[H]%
\centering
\subfigure[\label{fig:firstA00}]{\includegraphics[height=1.5in]{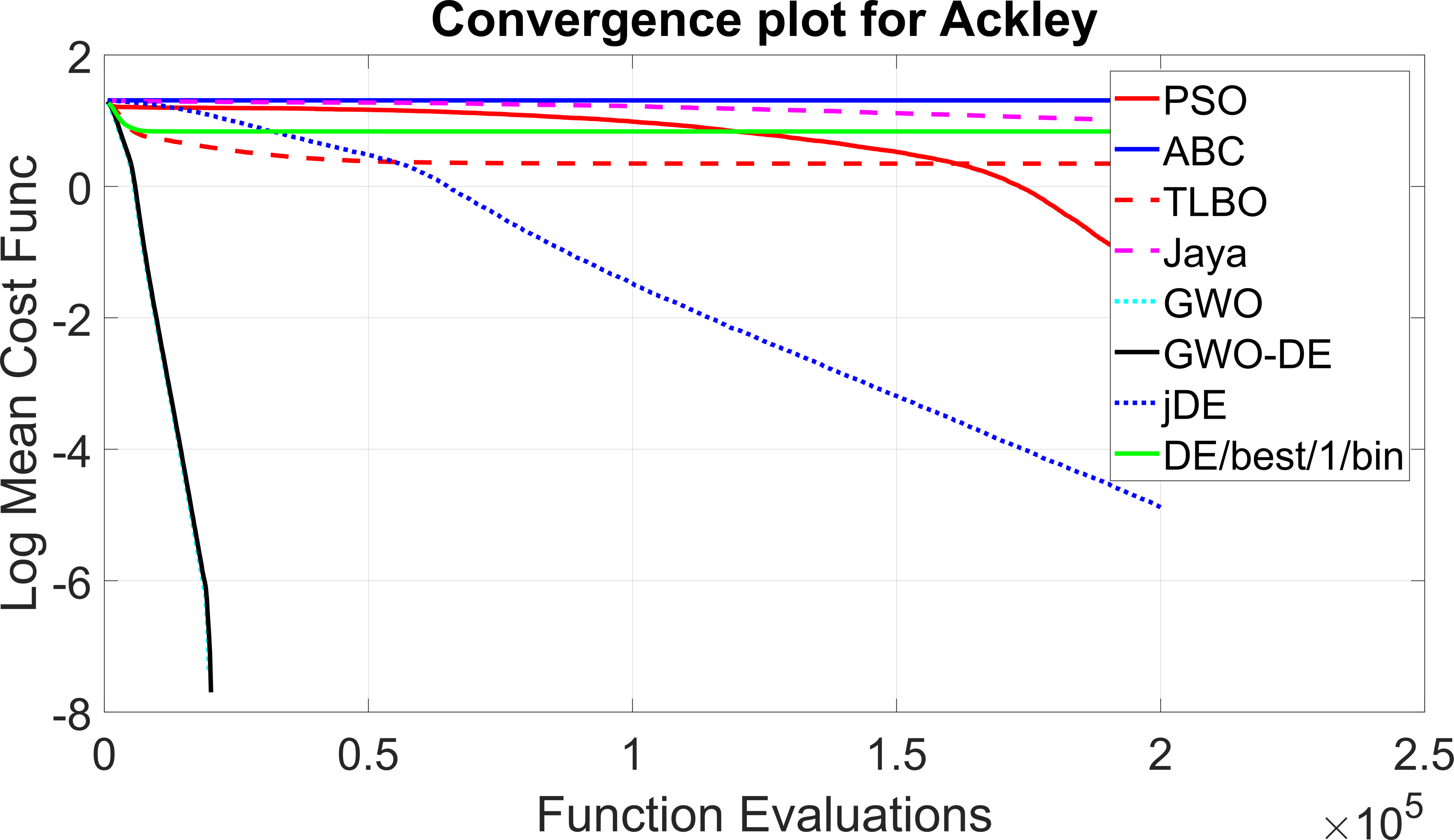}}%
\hfill
\subfigure[\label{fig:firstB00}]{\includegraphics[height=1.5in]{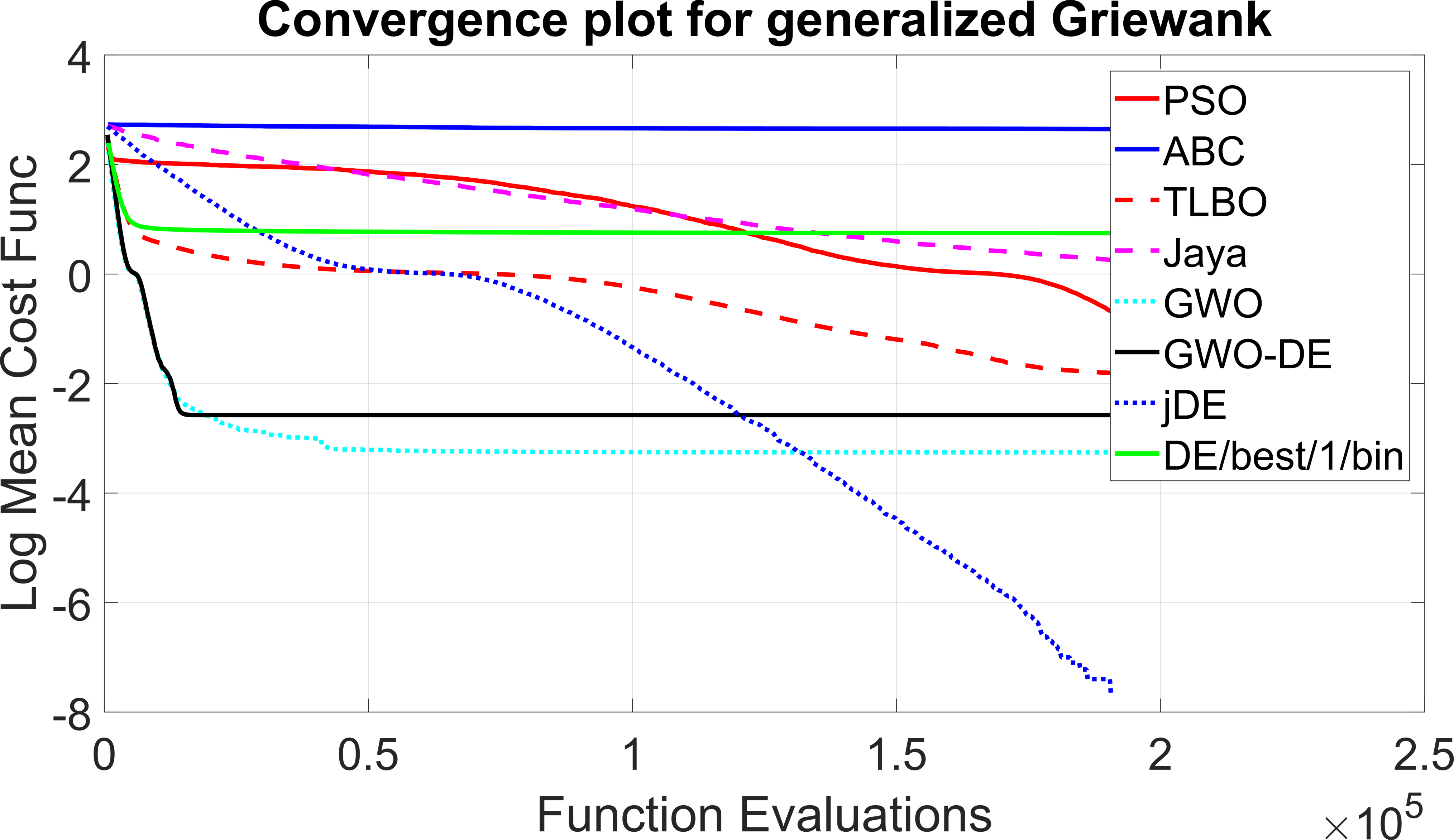}}%
\hfill
\subfigure[\label{fig:firstC00}]{\includegraphics[height=1.5in]{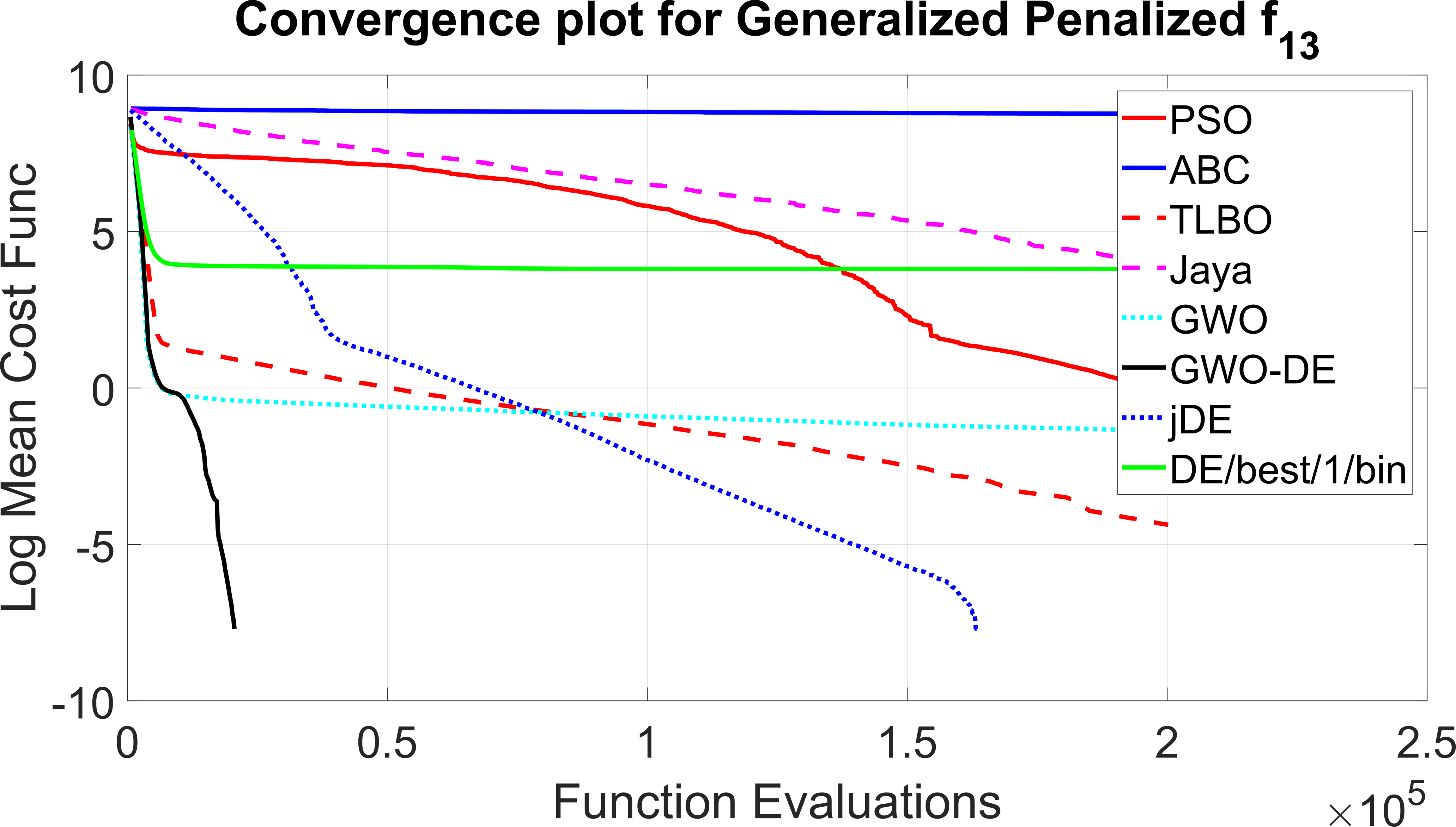}}%
\hfill
\subfigure[\label{fig:firstD00}]{\includegraphics[height=1.5in]{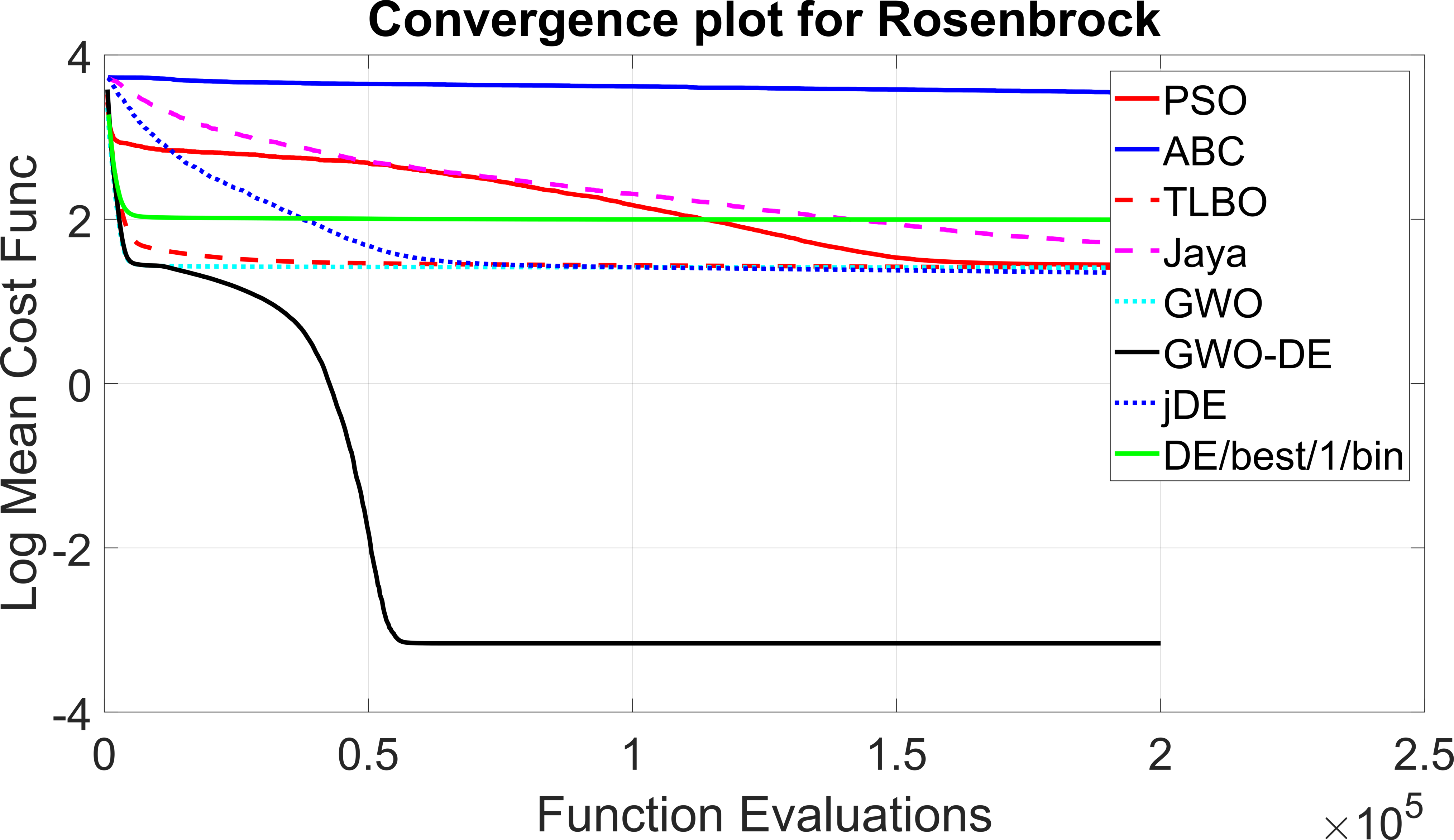}}%
\caption{Convergence rate plots of log mean error distribution for the common  benchmark functions, $D=30$, a) Ackley, b) generalized Griewank, c) Generalized Penalized $f_{13}$, d) Rosenbrock.}
\label{fig:convCTF30}%
\end{figure}

Figure \ref{fig:convCTF30} plots the average convergence rate graphs for the same four functions. We notice that GWO-DE obtains its final value only in about $20000$ function evaluations for the Ackley and Generalized Penalized $f_{13}$ case. For the generalized Griewank case, GWO-DE seems to converge faster than the other algorithms at the same speed as \emph{DE/best/1/bin}. Moreover, again for the Rosenbrock case, GWO-DE converges faster than the other algorithms at a lower value.

\subsection{CEC 2017 benchmark functions}

The CEC 2017 test suite contains 30 different test functions that belong to four distinct categories\cite{CEC2017}. All functions are minimization problems that are shifted to a global optimum $\vv{\sigma_k}=[\sigma_{1,k},\sigma_{2,k},\hdots,\sigma_{D,k}]$ which is randomly distributed in $[-80,80]$.
The search range for all functions is the same and is within the interval $[-100,100]$. The test suite contains unimodal functions ($f_1$,$f_2$), simple
multimodal functions ($f_3,\hdots, f_{10}$), hybrid functions ($f_{11},\hdots, f_{20}$), and composition functions ($f_{21},\hdots, f_{30}$).
The authors of \cite{CEC2017} have excluded $f_2$ because it shows unstable behavior. The hubrid functions divide the variables randomly into some subcomponents and then different basic functions are used for different subcomponents. The composition functions use a combination of basic functions. All the basic functions that have been used in composition functions are shifted and rotated functions.

In all cases, we compare GWO-DE with the same seven algorithms as used in the previous section. We apply the CEC 2017 test suite for three different cases with $D=30$, $D=50$, and $D=100$. In all cases we make the boxplots and convergence rate graphs for four distinct test functions, one of each category. These are namely $f_1$ (unimodal), $f_4$ (simple multimodal), $f_{11}$ (hybrid functions), and $f_{21}$ (composition functions).

\subsubsection{30 Dimensions}

\begin{figure}[H]%
\centering
\subfigure[\label{fig:firstA}]{\includegraphics[height=1.5in]{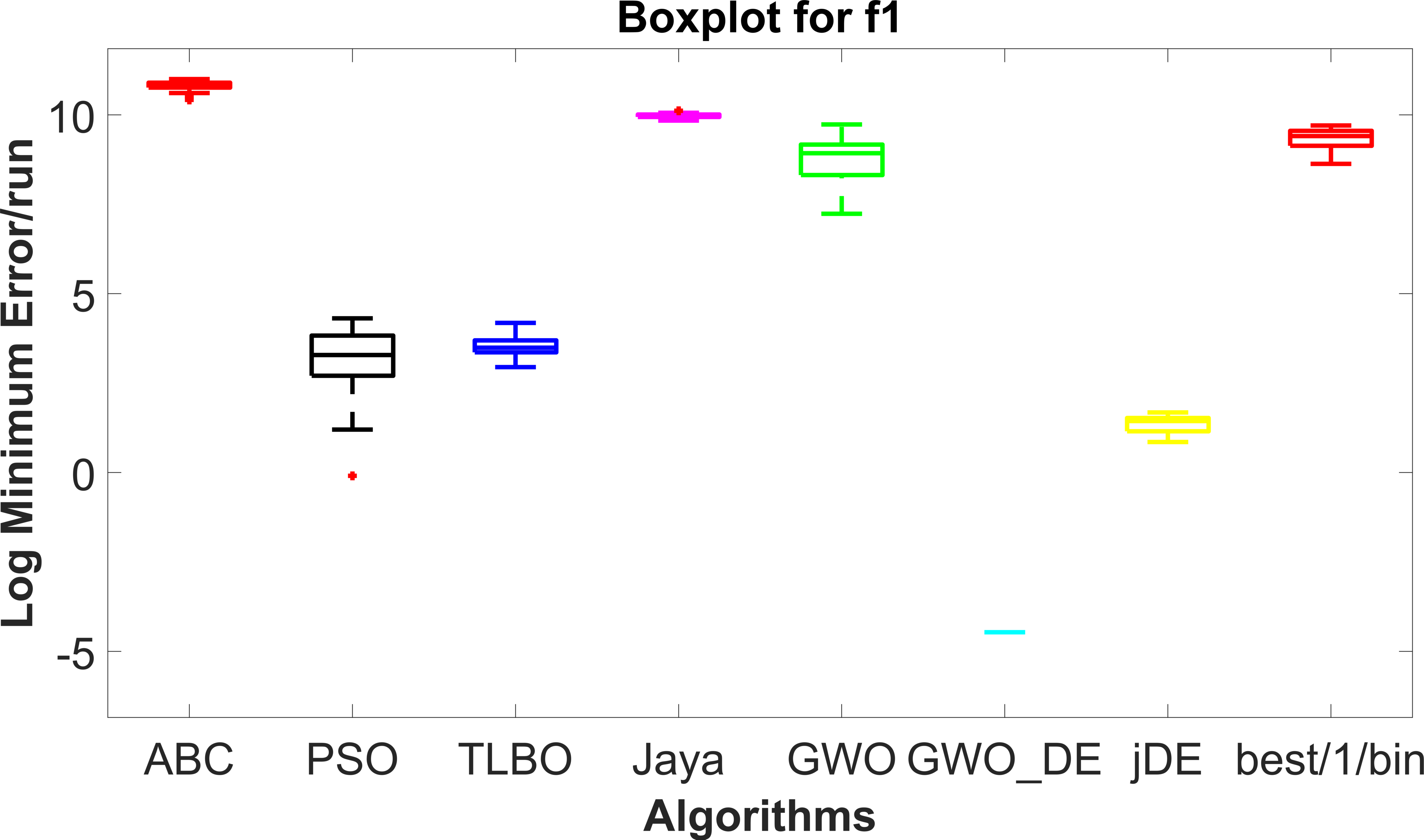}}%
\hfill
\subfigure[\label{fig:firstB}]{\includegraphics[height=1.5in]{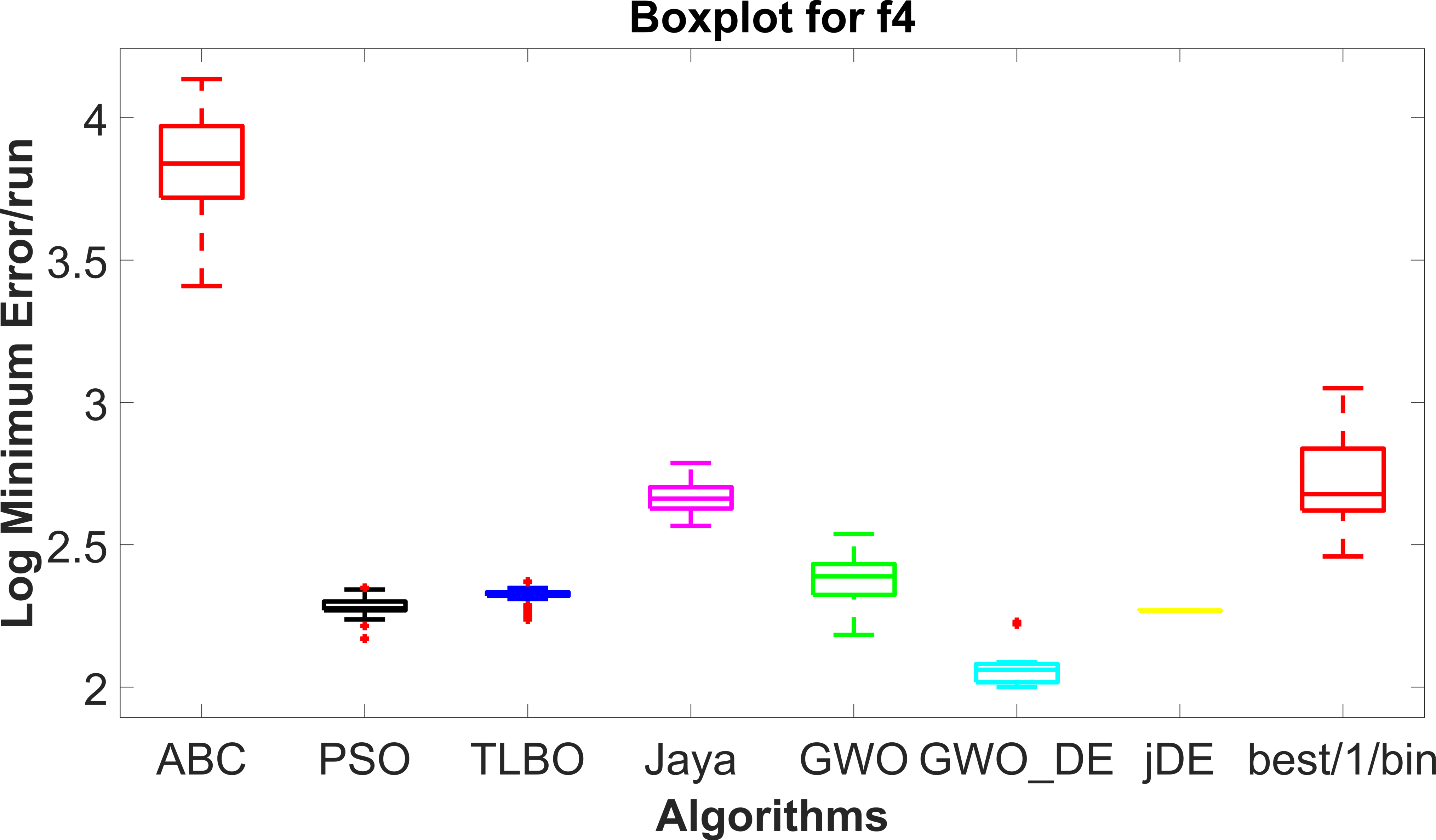}}%
\hfill
\subfigure[\label{fig:firstC}]{\includegraphics[height=1.5in]{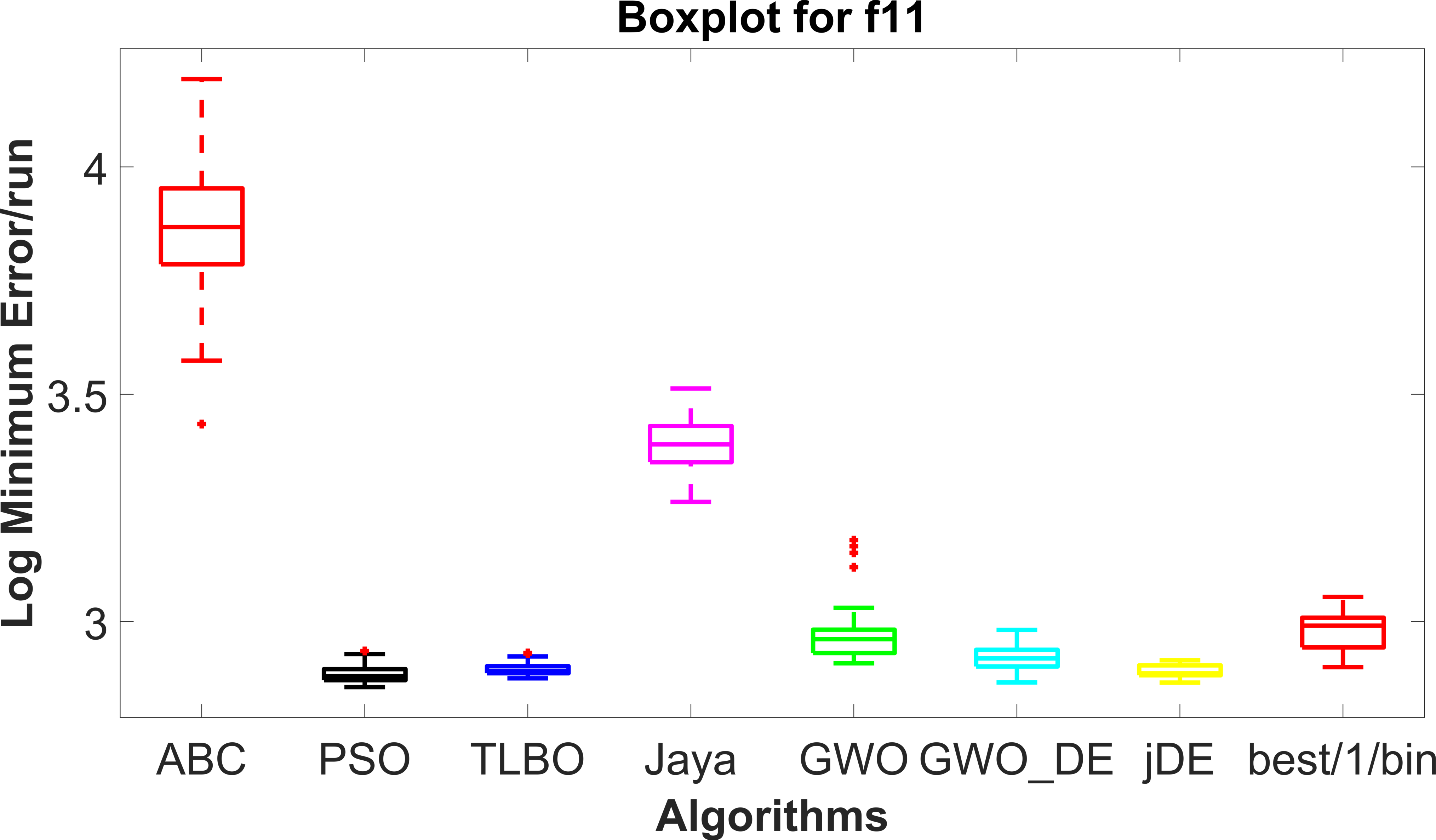}}%
\hfill
\subfigure[\label{fig:firstD}]{\includegraphics[height=1.5in]{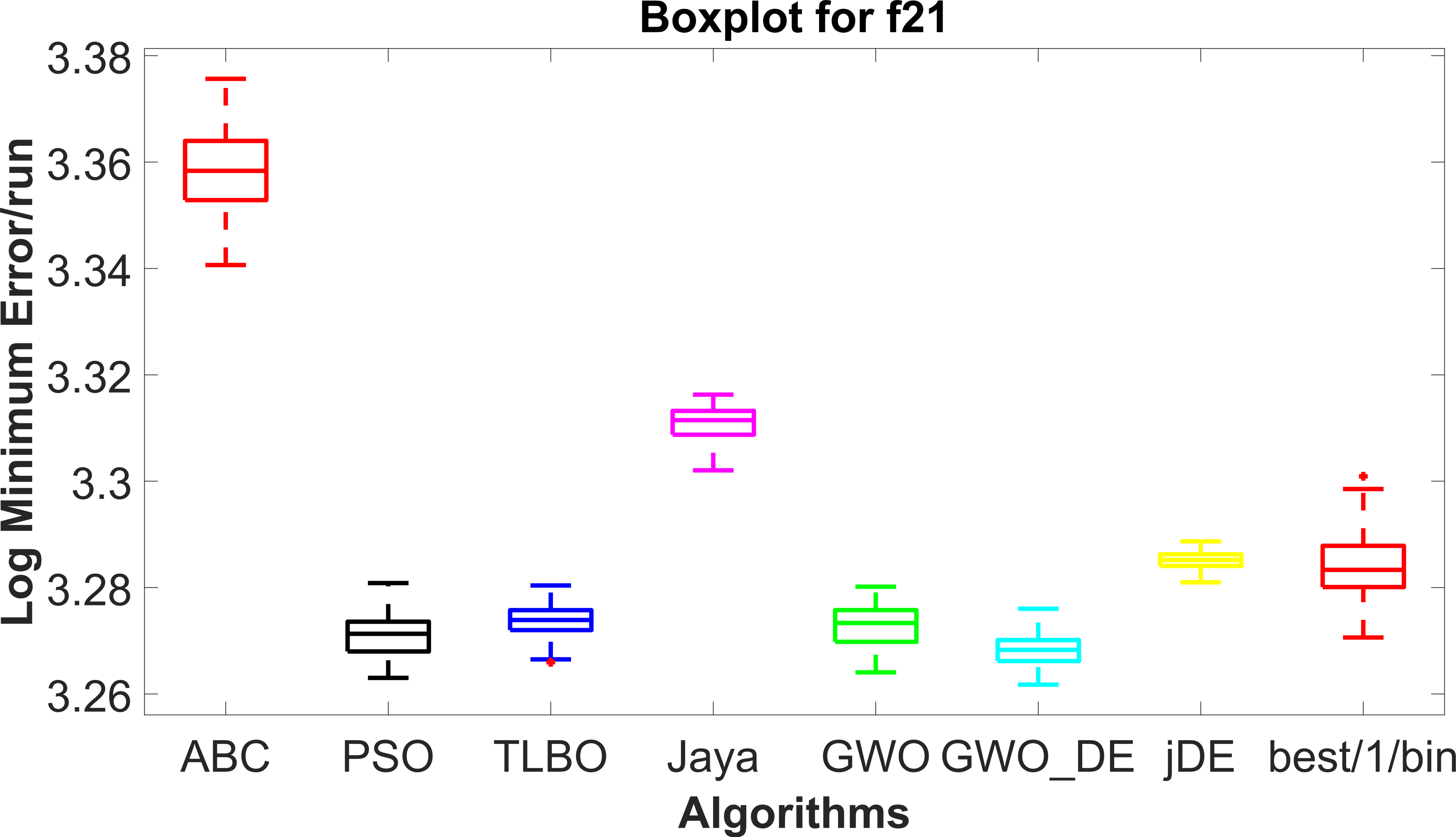}}%
\caption{Box plots of log error distribution for CEC 2017 benchmark functions, $D=30$, a) $f_1$, b) $f_4$, c) $f_{11}$, d) $f_{21}$.}
\label{fig:boxCEC30}%
\end{figure}

In this subsection, we run each algorithm independently $50$ times per case, with a population size equal to $200$ and a maximum number of objective function evaluations equal to $300,000$. Table \ref{tab1CEC30} holds the comparative results. From this Table it is evident that the specific reference functions are very demanding, since the average error values in the majority of cases do not approach very close to zero. However, it is observed that GWO-DE has the smallest error value in $10$ out of $29$ cases, more times than any other algorithm, while in $8$ cases it has the second smallest error.
Furthermore, the nonparametric test reported in Table \ref{tab2CEC30} shows that GWO-DE again ranks first in the Friedman test. In this case, PSO ranks second and TLBO third.
The distribution of the mean error per run is shown in Figure \ref{fig:boxCEC30} for four different benchmark functions.  We notice that GWO outperformed the other algorithms in terms of smaller values and smaller distribution. One may notice that GWO-DE performed significantly better than the three algorithms composed of namely GWO, jDE and \emph{DE/best/1/bin}.

\begin{table}[h]
\caption{Comparative results of CEC 2017 benchmark functions for $D=30$. The bold font indicates the smaller value.\label{tab1CEC30}}
\newcolumntype{C}{>{\centering\arraybackslash}X}
\begin{tabularx}{\textwidth}{|C|C|C|C|C|C|C|C|C|}
\hline\hline
\textbf{Test   function}     & \textbf{ABC} & \textbf{PSO}      & \textbf{TLBO} & \textbf{Jaya} & \textbf{GWO}      & \textbf{GWO-DE}   & \textbf{jDE}      & \textbf{{DE/
best/
1/
bin}} \\
\hline
$f_1$                              & 6.82
E+13                                          & 4.50
E+03                                          & 4.17E+03                                           & 9.41
E+09                                           & 1.07
E+09                                          & \textbf{7.00
E-07}                                    & 2.59
E+01                                          & 2.50
E+09                                                    \\\hline
$f_3$                              & 1.24
E+05                                          & 1.51
E+04                                          & 4.44
E+04                                           & 9.72
E+04                                           & 2.84
E+04                                          & 3.14
E+03                                             & 4.36
E+04                                          & \textbf{7.83
E-04}                                           \\\hline
$f_4$                              & 7.18
E+03                                          & 9.31
E+01                                          & 1.10
E+02                                           & 3.67
E+02                                           & 1.48
E+02                                          & \textbf{1.46
E+01}                                    & 8.57
E+01                                          & 4.36
E+02                                                    \\\hline
$f_5$                              & 5.55
E+02                                          & 7.07
E+01                                          & 9.63
E+01                                           & 2.64
E+02                                           & 8.98
E+01                                          & \textbf{6.59
E+01}                                    & 1.27
E+02                                          & 1.32
E+02                                                    \\\hline
$f_6$                              & 1.12
E+02                                          & 1.49
E-01                                          & 3.41
E-01                                           & 3.43
E+01                                           & 4.58
E+00                                          & 5.19
E+00                                             & \textbf{8.40
E-06}                                 & 1.97
E+01                                                    \\\hline
$f_7$                              & 2.11
E+03                                          & 1.20
E+02                                          & 1.15
E+02                                           & 4.04
E+02                                           & 1.36
E+02                                          & \textbf{1.07
E+02}                                    & 1.61
E+02                                          & 2.06
E+02                                                    \\\hline
$f_8$                              & 5.00
E+02                                          & \textbf{6.25
E+01}                                 & 7.53
E+01                                           & 2.70
E+02                                           & 7.77
E+01                                          & 6.40
E+01                                             & 1.31
E+02                                          & 1.07
E+02                                                    \\\hline
$f_9$                              & 1.42
E+04                                          & 1.24
E+02                                          & 2.26
E+01                                           & 5.52
E+03                                           & 5.00
E+02                                          & 4.38
E+02                                             & \textbf{0.00
E+00}                                 & 1.92
E+03                                                    \\\hline
$f_{10}$                           & 7.61
E+03                                          & 3.48
E+03                                          & 6.80
E+03                                           & 6.82
E+03                                           & 3.27
E+03                                          & \textbf{2.97
E+03}                                    & 5.27
E+03                                          & 3.46
E+03                                                    \\\hline

\end{tabularx}
\vspace{5pt}
\end{table}
\FloatBarrier
\begin{table}[h]
\newcolumntype{C}{>{\centering\arraybackslash}X}
\begin{tabularx}{\textwidth}{|C|C|C|C|C|C|C|C|C|}
\hline
$f_{11}$                           & 7.11
E+03                                          & \textbf{6.73
E+01}                                 & 8.72
E+01                                           & 1.79
E+03                                           & 2.49
E+02                                          & 1.36
E+02                                             & 7.88
E+01                                          & 2.60
E+02                                                    \\\hline
$f_{12}$                           & 8.69
E+09                                          & 6.09
E+05                                          & \textbf{3.23
E+05}                                  & 1.84
E+08                                           & 4.12
E+07                                          & 1.17
E+06                                             & 3.96
E+05                                          & 6.10
E+07                                                    \\\hline
$f_{13}$                           & 6.25
E+09                                          & 4.40
E+04                                          & 9.43
E+03                                           & 2.57
E+07                                           & 1.39
E+05                                          & 5.81
E+03                                             & \textbf{6.00
E+02}                                 & 1.68
E+05                                                    \\\hline
$f_{14}$                           & 2.07
E+06                                          & 4.28
E+04                                          & 3.48
E+03                                           & 1.56
E+05                                           & 1.14
E+05                                          & 2.58
E+02                                             & \textbf{7.13
E+01}                                 & 9.42
E+02                                                    \\\hline
$f_{15}$                           & 5.76
E+08                                          & 1.44
E+04                                          & 6.78
E+02                                           & 1.12
E+07                                           & 2.60
E+05                                          & 7.28
E+02                                             & \textbf{6.67
E+01}                                 & 1.38
E+04                                                    \\\hline
$f_{16}$                           & 2.57
E+03                                          & 7.01
E+02                                          & 7.65
E+02                                           & 1.75
E+03                                           & 6.97
E+02                                          & \textbf{5.71
E+02}                                    & 7.64
E+02                                          & 9.58
E+02                                                    \\\hline
$f_{17}$                           & 1.36
E+03                                          & 2.51
E+02                                          & \textbf{1.82
E+02}                                  & 7.03
E+02                                           & 2.61
E+02                                          & 2.26
E+02                                             & 2.30
E+02                                          & 5.42
E+02                                                    \\\hline
$f_{18}$                           & 7.64
E+06                                          & 4.13
E+05                                          & 1.85
E+05                                           & 2.33
E+06                                           & 9.43
E+05                                          & 2.69
E+04                                             & \textbf{1.44
E+03}                                 & 2.71
E+03                                                    \\\hline
$f_{19}$                           & 5.67
E+08                                          & 8.35
E+03                                          & 3.50
E+03                                           & 1.02
E+06                                           & 2.82
E+05                                          & 2.67
E+02                                             & \textbf{3.54
E+01}                                 & 3.32
E+04                                                    \\\hline
$f_{20}$                           & 1.27
E+03                                          & \textbf{1.50
E+02}                                 & 2.33
E+02                                           & 6.21
E+02                                           & 3.09
E+02                                          & 3.40
E+02                                             & 1.88
E+02                                          & 4.52
E+02                                                    \\\hline
$f_{21}$                           & 6.83
E+02                                          & 2.66
E+02                                          & 2.79
E+02                                           & 4.46
E+02                                           & 2.75
E+02                                          & \textbf{2.56
E+02}                                    & 3.28
E+02                                          & 3.25
E+02                                                    \\\hline
$f_{22}$                           & 7.43
E+03                                          & 7.11
E+02                                          & \textbf{1.01
E+02}                                  & 5.00
E+03                                           & 2.27
E+03                                          & 1.88
E+03                                             & 2.12
E+02                                          & 2.52
E+03                                                    \\\hline
$f_{23}$                           & 1.09
E+03                                          & 4.25
E+02                                          & \textbf{4.22
E+02}                                  & 6.17
E+02                                           & 4.48
E+02                                          & 4.30
E+02                                             & 4.77
E+02                                          & 5.52
E+02                                                    \\\hline
$f_{24}$                           & 1.27
E+03                                          & 5.05
E+02                                          & 4.98
E+02                                           & 6.69
E+02                                           & 5.40
E+02                                          & \textbf{4.98
E+02}                                    & 5.59
E+02                                          & 6.32
E+02                                                    \\\hline
$f_{25}$                           & 2.52
E+03                                          & 3.87
E+02                                          & 4.10
E+02                                           & 5.45
E+02                                           & 4.38
E+02                                          & \textbf{3.80
E+02}                                    & 3.87
E+02                                          & 5.75
E+02                                                    \\\hline
$f_{26}$                           & 7.52
E+03                                          & 1.67
E+03                                          & 1.19
E+03                                           & 3.98
E+03                                           & \textbf{4.52
E+02}                                 & 7.63
E+02                                             & 2.26
E+03                                          & 3.27
E+03                                                    \\\hline
$f_{27}$                           & 1.00
E+03                                          & 5.16
E+02                                          & 5.22
E+02                                           & 5.47
E+02                                           & \textbf{5.00
E+02}                                 & 5.00
E+02                                             & 5.00E+02                                          & 5.89
E+02                                                    \\\hline
$f_{28}$                           & 3.54
E+03                                          & 4.26
E+02                                          & 4.69
E+02                                           & 1.14
E+03                                           & 2.39
E+02                                          & \textbf{2.22
E+02}                                    & 4.07
E+02                                          & 8.14
E+02                                                    \\\hline
$f_{29}$                           & 2.14
E+03                                          & 6.60
E+02                                          & 5.84
E+02                                           & 1.47
E+03                                           & \textbf{5.40
E+02}                                 & 6.43
E+02                                             & 7.58
E+02                                          & 1.09
E+03                                                    \\\hline
$f_{30}$                           & 1.20
E+08                                          & 1.08
E+04                                          & \textbf{3.83
E+03}                                  & 1.67
E+07                                           & 2.34
E+05                                          & 4.44
E+04                                             & 4.68
E+03                                          & 9.86
E+05    \\
\hline\hline
\end{tabularx}
\vspace{5pt}
\end{table}
\FloatBarrier
\begin{table}[h]
\caption{Average Rankings achieved by Friedman test for CEC 2017 benchmark functions for $D=30$. The bold font indicates the smaller value.\label{tab2CEC30}}
\newcolumntype{C}{>{\centering\arraybackslash}X}
\begin{tabularx}{\textwidth}{|C|C|C|C|}
\hline\hline
\textbf{Algorithm}	& \textbf{Average Ranking}	& \textbf{Normalized} & \textbf{Ranks}\\\hline
ABC           & 8.00             & 3.42       & 8          \\ \hline
PSO           & 3.17          & 1.35       & 4          \\ \hline
TLBO          & 3.10           & 1.32       & 3          \\ \hline
Jaya          & 6.90           & 2.95       & 7          \\ \hline
GWO           & 4.03          & 1.72       & 5          \\ \hline
GWO-DE        & \textbf{2.34} & \textbf{1.00} & \textbf{1} \\ \hline
jDE           & 3.03          & 1.29       & 2          \\ \hline
\emph{DE/best/1/bin} & 5.41          & 2.31       & 6    \\
\hline\hline
\end{tabularx}
\vspace{5pt}
\end{table}
\FloatBarrier
Regarding the convergence speed of GWO-DE, this was examined through the convergence plots drawn separately for each case. The graphs of four functions are illustrated in Figure \ref{fig:convCEC30}. GWO-DE is found to converge to the final solution faster than its competitors in most cases. Additionally, GWO-DE requires fewer objective function evaluations to converge to the final value than all the other algorithms.
\begin{figure}[H]%
\centering
\subfigure[\label{fig:firstA}]{\includegraphics[height=1.5in]{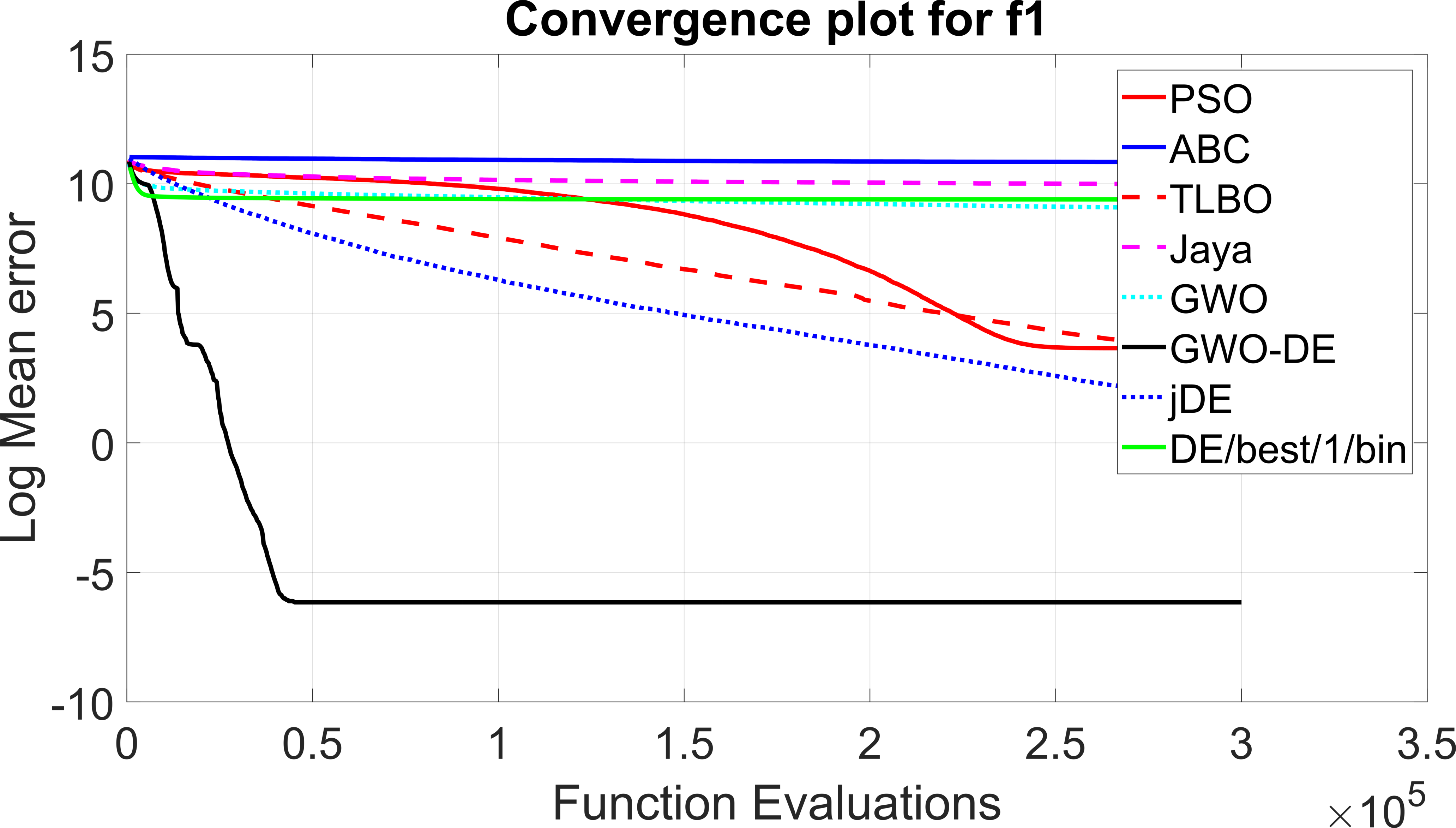}}%
\hfill
\subfigure[\label{fig:firstB}]{\includegraphics[height=1.5in]{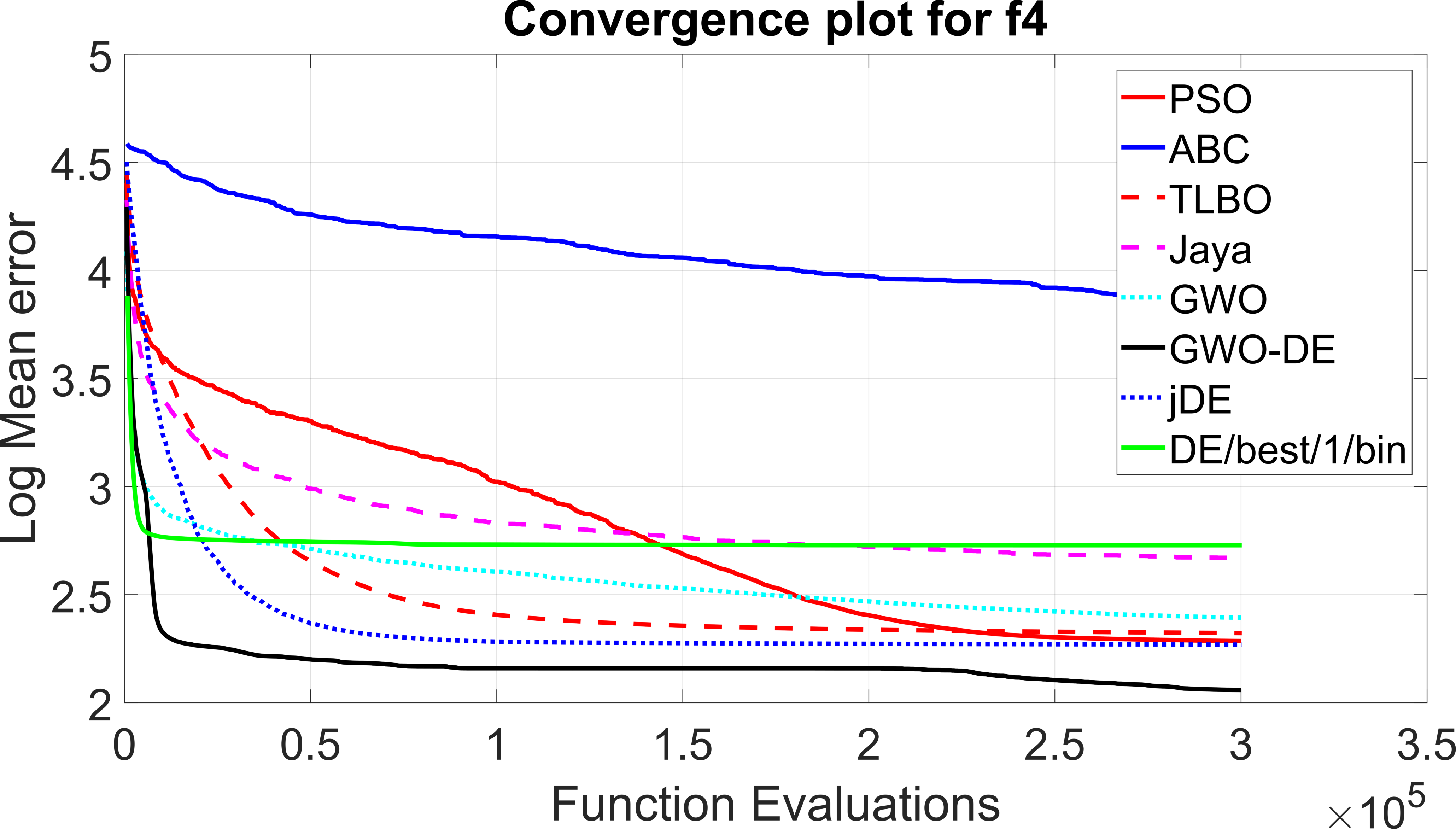}}%
\hfill
\subfigure[\label{fig:firstC}]{\includegraphics[height=1.5in]{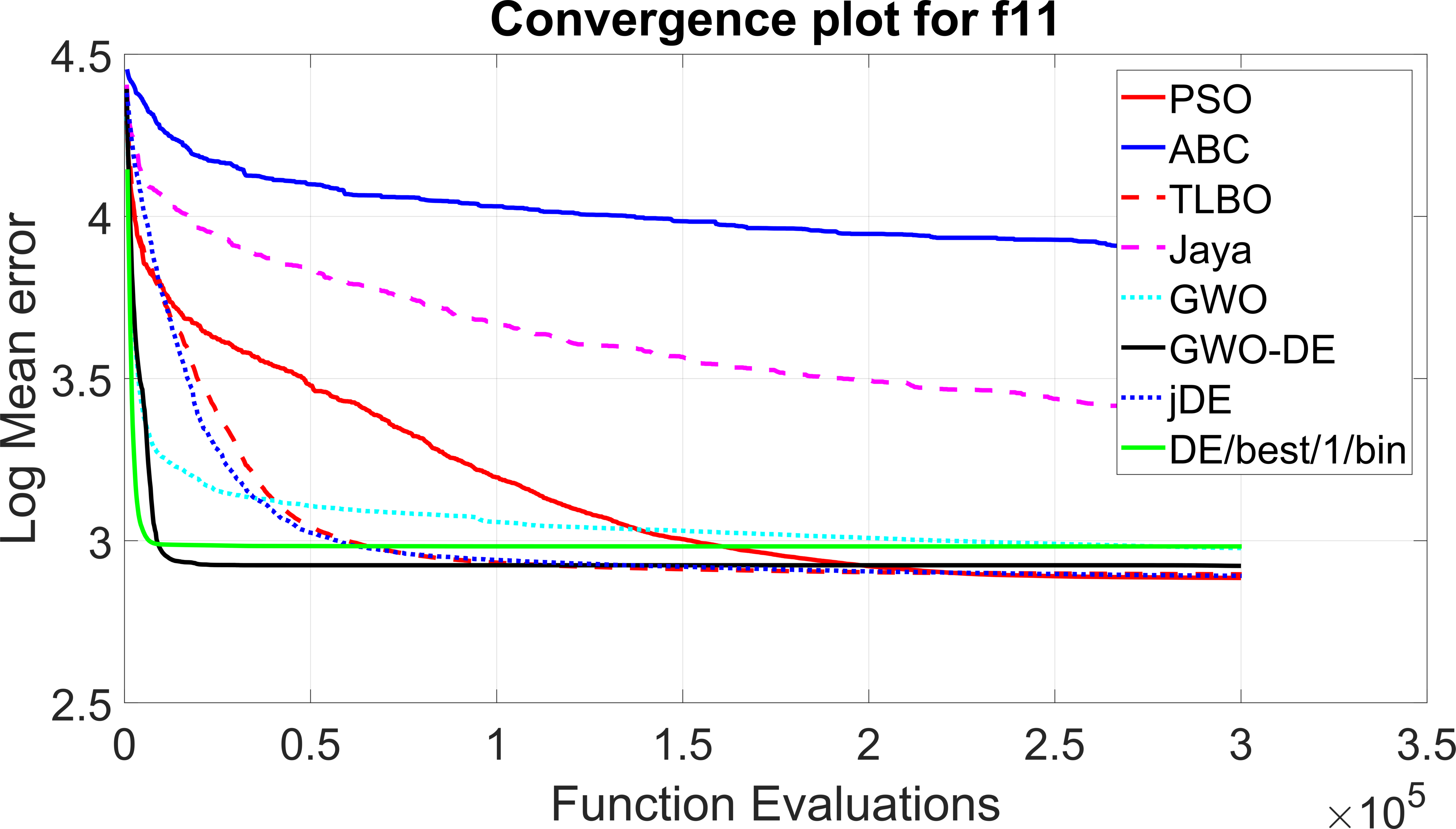}}%
\hfill
\subfigure[\label{fig:firstD}]{\includegraphics[height=1.5in]{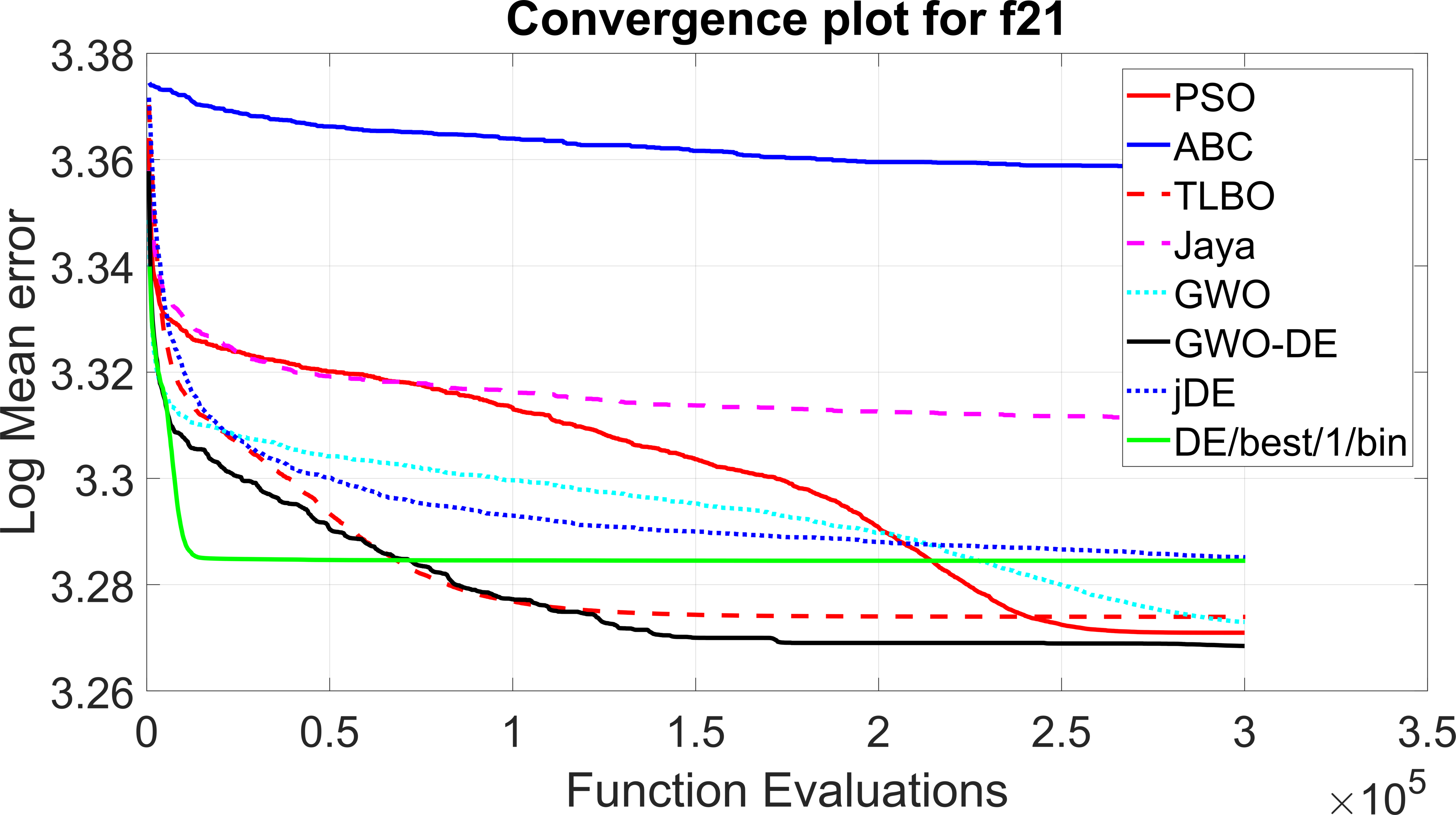}}%
\caption{Convergence rate plots of log mean error distribution for CEC 2017 benchmark functions, $D=30$, a) $f_1$, b) $f_4$, c) $f_{11}$, d) $f_{21}$.}
\label{fig:convCEC30}%
\end{figure}

\subsubsection{50 Dimensions}

\begin{figure}[H]%
\centering
\subfigure[\label{fig:firstA3}]{\includegraphics[height=1.5in]{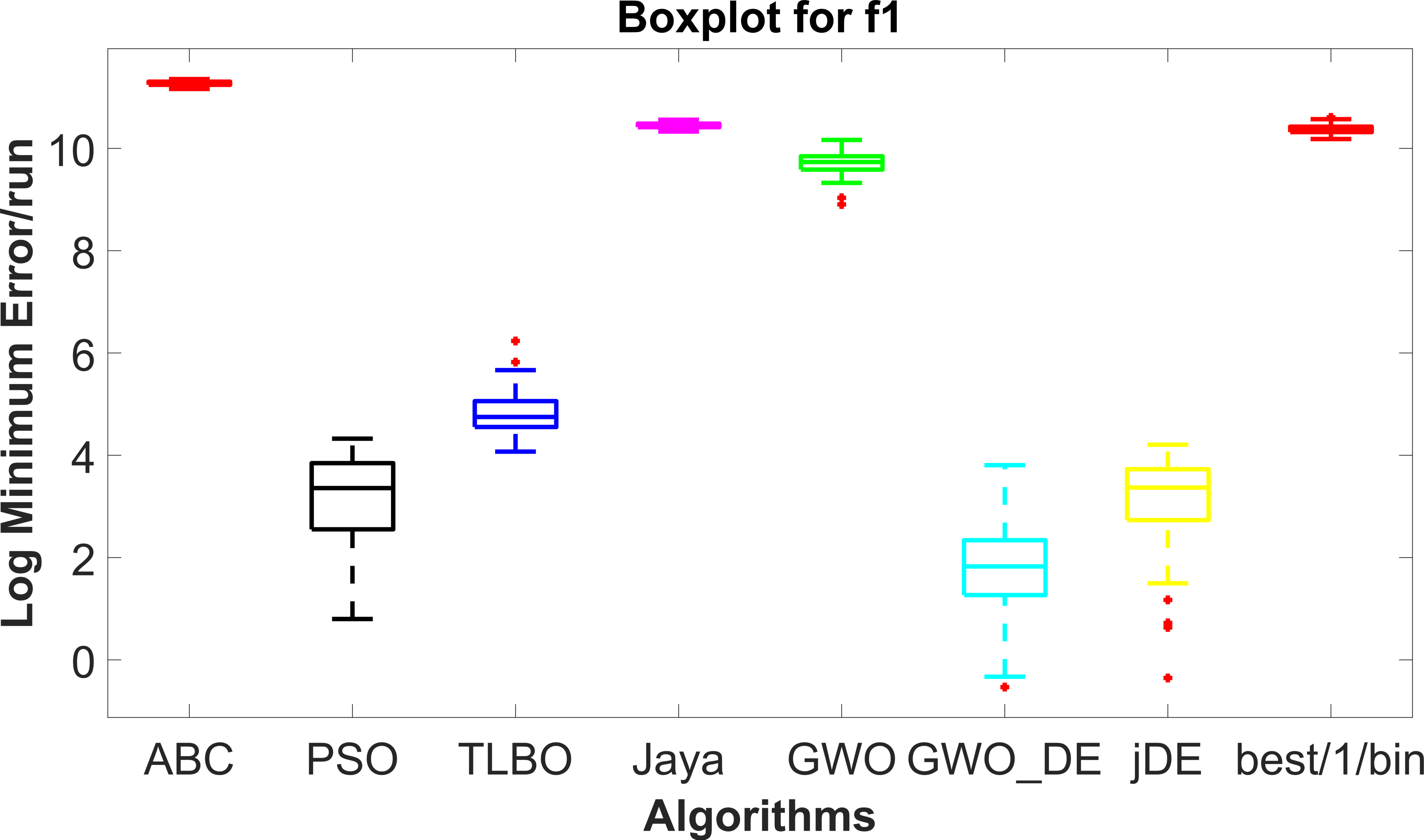}}%
\hfill
\subfigure[\label{fig:firstB3}]{\includegraphics[height=1.5in]{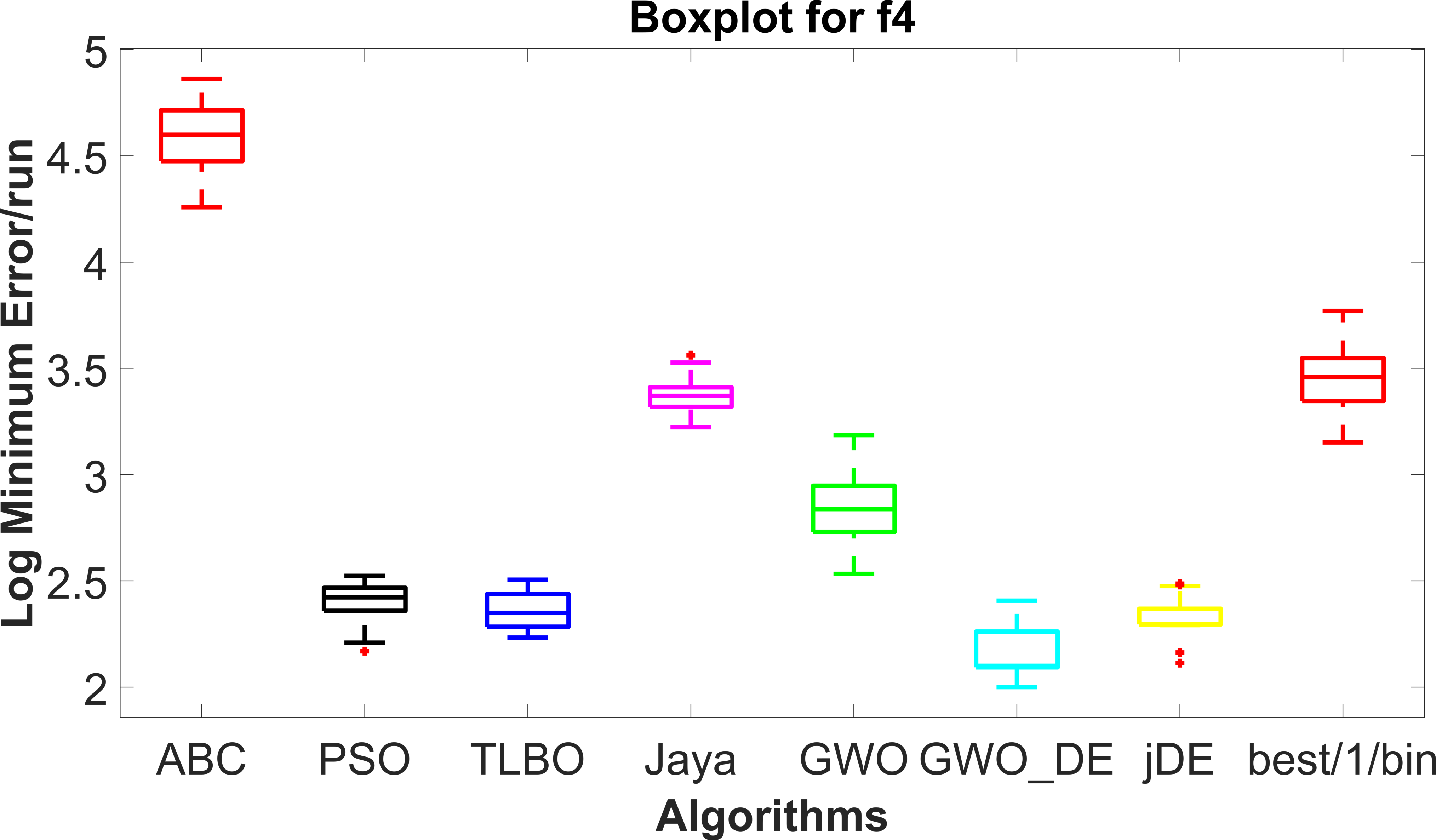}}%
\hfill
\subfigure[\label{fig:firstC3}]{\includegraphics[height=1.5in]{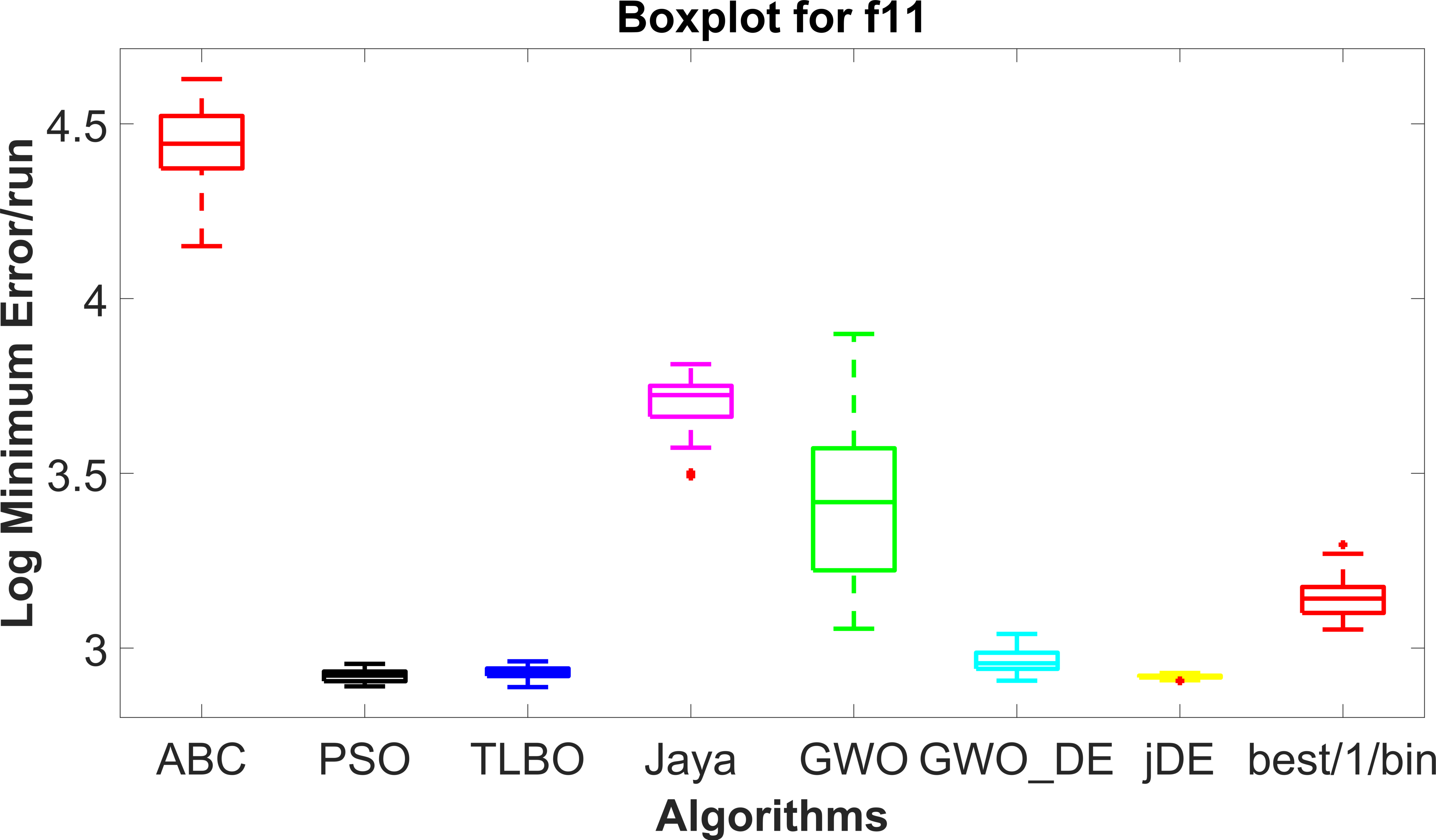}}%
\hfill
\subfigure[\label{fig:firstD3}]{\includegraphics[height=1.5in]{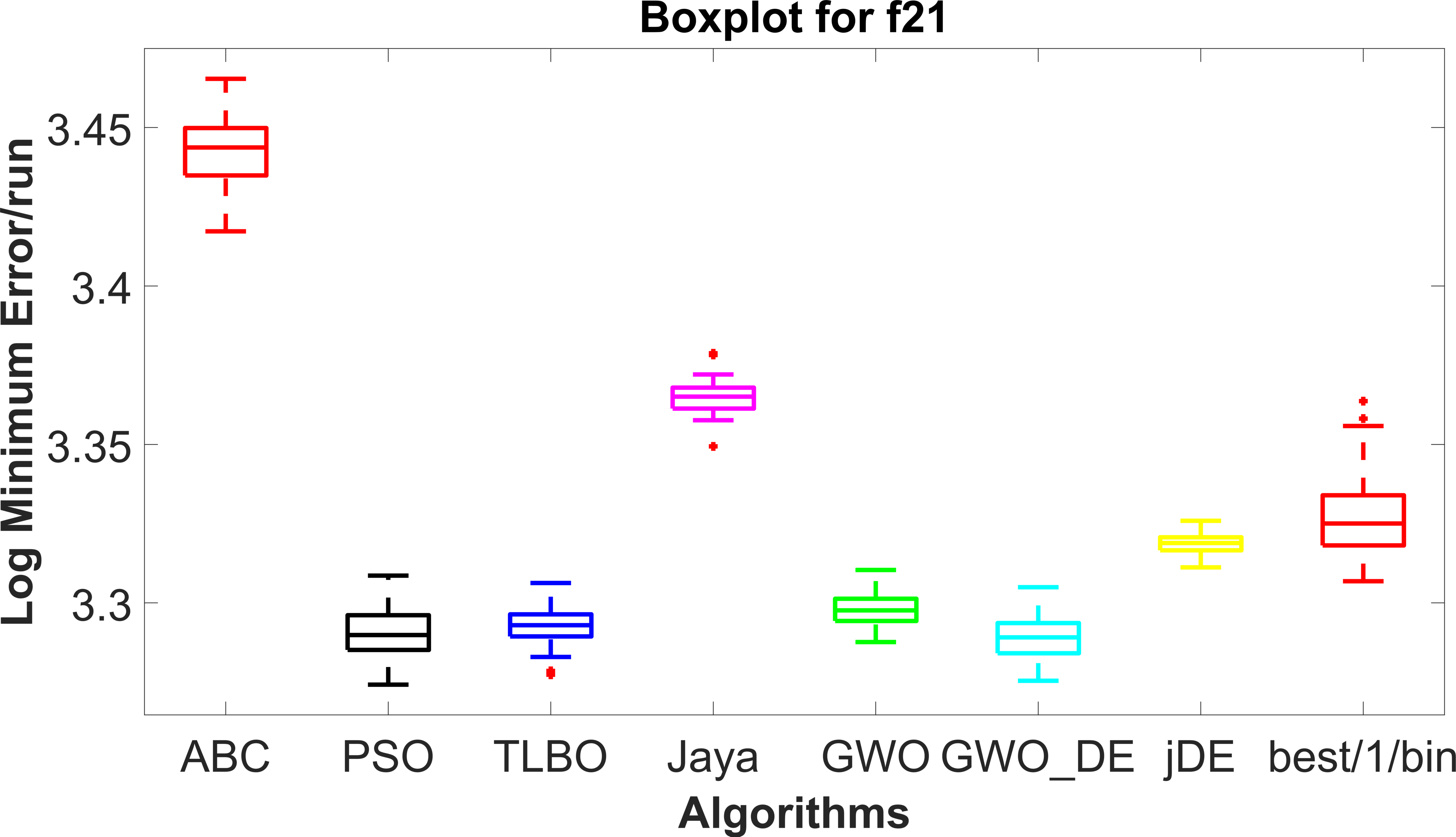}}%
\caption{Box plots of log error distribution for CEC 2017 benchmark functions, $D=50$, a) $f_1$, b) $f_4$, c) $f_{11}$, d) $f_{21}$.}
\label{fig:boxCEC50}%
\end{figure}

The next results present the same CEC 2017 benchmark problems in $50$ dimensions. Thus, the difficulty level is higher. Again, each algorithm was run independently $50$ times per case, with a population size equal to $200$ and a maximum number of objective function evaluations equal to $500,000$. Table \ref{tab1CEC50} lists the comparative mean error values of all algorithms.
We can observe that GWO-DE shows the smallest mean error value in $6$ out of the $28$ problems, while jDE in $8$. However, if we also consider the second smallest mean error value, then GWO-DE has the best or second best performance on $17$ problems in total, more times than any other algorithm.
Furthermore, as we see from Table \ref{tab2CEC50}
Still, it again ranks first in the Friedman test ranking and is significantly better than ABC, Jaya, GWO, and \emph{DE/best/1/bin}.
\begin{table}[h]
\caption{Comparative results of CEC 2017 benchmark functions for $D=50$. The bold font indicates the smaller value.\label{tab1CEC50}}
\newcolumntype{C}{>{\centering\arraybackslash}X}
\begin{tabularx}{\textwidth}{|C|C|C|C|C|C|C|C|C|}
\hline\hline
\textbf{Test   function}     & \textbf{ABC} & \textbf{PSO}      & \textbf{TLBO} & \textbf{Jaya} & \textbf{GWO}      & \textbf{GWO-DE}   & \textbf{jDE}      & \textbf{{DE/
best/
1/
bin}} \\
\hline
$f_1$                              & 1.88
E+11                      & 4.68
E+03                      & 1.26
E+05                       & 2.82
E+10                       & 5.78
E+09                      & \textbf{3.62
E+02}                & 3.54
E+03                      & 2.43
E+10                                \\\hline
$f_3$                              & 2.39
E+05                      & 5.55
E+04                      & 1.10
E+05                       & 2.05
E+05                       & 7.77
E+04                      & 3.46
E+04                         & 1.96
E+05                      & \textbf{6.00
E-07}                       \\\hline
$f_4$                              & 4.06
E+04                      & 1.60
E+02                      & 1.34
E+02                       & 2.29
E+03                       & 6.40
E+02                      & \textbf{4.66
E+01}                & 1.15
E+02                      & 2.94
E+03                                \\\hline
$f_5$                              & 1.01
E+03                      & \textbf{1.54
E+02}             & 2.06
E+02                       & 5.31
E+02                       & 1.89
E+02                      & \textbf{1.55
E+02}                & 2.77
E+02                      & 3.09
E+02                                \\\hline
$f_6$                              & 1.33
E+02                      & 1.69
E+00                      & 3.76
E+00                       & 5.12
E+01                       & 1.33
E+01                      & 1.51
E+01                         & \textbf{1.05
E-05}             & 4.13
E+01                                \\\hline
$f_7$                              & 4.53
E+03                      & 2.86
E+02                      & 2.46
E+02                       & 8.08
E+02                       & 2.85
E+02                      & \textbf{2.38
E+02}                & 3.43
E+02                      & 7.25
E+02                                \\\hline
$f_8$                              & 1.04
E+03                      & \textbf{1.47
E+02}             & 2.11
E+02                       & 5.66
E+02                       & 2.04
E+02                      & 1.61
E+02                         & 2.79
E+02                      & 2.96
E+02                                \\\hline
$f_9$                              & 5.15
E+04                      & 1.90
E+03                      & 3.08
E+02                       & 1.56
E+04                       & 5.09
E+03                      & 2.50
E+03                         & \textbf{1.00
E-07}             & 8.33
E+03                                \\\hline
$f_{10}$                           & 1.37
E+04                      & 5.86
E+03                      & 1.04
E+04                       & 1.33
E+04                       & 6.24
E+03                      & \textbf{5.24
E+03}                & 1.02
E+04                      & 6.63
E+03                                \\\hline
$f_{11}$                           & 2.76
E+04                      & \textbf{1.32
E+02}             & 1.50
E+02                       & 4.45
E+03                       & 2.18
E+03                      & 2.21
E+02                         & \textbf{1.28
E+02}             & 7.08
E+02                                \\\hline
$f_{12}$                           & 7.71
E+10                      & 2.53
E+06                      & \textbf{1.29
E+06}              & 4.99
E+09                       & 6.23
E+08                      & \textbf{7.84
E+05}                & 1.60
E+06                      & 1.49
E+09                                \\\hline
$f_{13}$                           & 3.71
E+10                      & 1.99
E+04                      & \textbf{1.66
E+03}              & 8.45
E+08                       & 1.28
E+08                      & 8.26
E+03                         & \textbf{6.25
E+03}             & 1.76
E+07                                \\\hline
$f_{14}$                           & 2.50
E+07                      & 1.21
E+05                      & 3.18
E+04                       & 1.18
E+06                       & 6.58
E+05                      & 5.44
E+04                         & \textbf{1.55
E+02}             & 9.47
E+03                                \\\hline
$f_{15}$                           & 1.17
E+10                      & 1.46
E+04                      & 1.05
E+04                       & 2.39
E+08                       & 6.30
E+06                      & 7.85
E+04                         & \textbf{2.20
E+02}             & 1.04
E+05                                \\\hline
$f_{16}$                           & 6.31
E+03                      & 1.41
E+03                      & \textbf{1.04
E+03}              & 3.62
E+03                       & 1.32
E+03                      & \textbf{1.33
E+03}                & 2.02
E+03                      & 1.86
E+03                                \\\hline
$f_{17}$                           & 1.77
E+04                      & 1.04
E+03                      & \textbf{1.06
E+03}              & 2.56
E+03                       & \textbf{9.33
E+02}             & 9.64
E+02                         & 1.13
E+03                      & 1.47
E+03                                \\\hline
$f_{18}$                           & 5.95
E+07                      & 1.50
E+06                      & 7.78
E+05                       & 1.31
E+07                       & 3.36
E+06                      & 9.64
E+04                         & \textbf{2.06
E+04}             & \textbf{6.28
E+03}                       \\\hline
\end{tabularx}
\vspace{5pt}
\end{table}
\FloatBarrier
\begin{table}[h]
\newcolumntype{C}{>{\centering\arraybackslash}X}
\begin{tabularx}{\textwidth}{|C|C|C|C|C|C|C|C|C|}
\hline

$f_{19}$                           & 3.19
E+09                      & 1.96
E+04                      & 1.52
E+04                       & 6.03
E+07                       & 1.32
E+06                      & 8.22
E+03                         & \textbf{8.66
E+01}             & 1.96
E+06                                \\\hline
$f_{20}$                           & 2.57
E+03                      & \textbf{7.54
E+02}             & \textbf{4.86
E+02}              & 1.82
E+03                       & 7.71
E+02                      & 7.69
E+02                         & 1.06
E+03                      & 1.01
E+03                                \\\hline
$f_{21}$                           & 1.17
E+03                      & 3.50
E+02                      & 3.63
E+02                       & 7.18
E+02                       & 3.86
E+02                      & \textbf{3.47
E+02}                & 4.82
E+02                      & 5.25
E+02                                \\\hline
$f_{22}$                           & 1.36
E+04                      & 7.64
E+03                      & \textbf{3.99
E+02}              & 1.34
E+04                       & 7.50
E+03                      & 6.31
E+03                         & 1.05
E+04                      & 7.29
E+03                                \\\hline
$f_{23}$                           & 1.93
E+03                      & \textbf{5.89
E+02}             & \textbf{6.17
E+02}              & 1.02
E+03                       & 6.44
E+02                      & 6.18
E+02                         & 7.09
E+02                      & 1.01
E+03                                \\\hline
$f_{24}$                           & 2.12
E+03                      & \textbf{6.57
E+02}             & 6.98
E+02                       & 1.01
E+03                       & 7.90
E+02                      & \textbf{6.84
E+02}                & 7.93
E+02                      & 1.05
E+03                                \\\hline
$f_{25}$                           & 1.50
E+04                      & 5.58
E+02                      & 6.35
E+02                       & 1.62
E+03                       & 8.71
E+02                      & \textbf{4.94
E+02}                & \textbf{4.82
E+02}             & 2.53
E+03                                \\\hline
$f_{26}$                           & 1.49
E+04                      & 2.97
E+03                      & 6.07
E+03                       & 7.46
E+03                       & \textbf{9.30
E+02}             & 2.30
E+03                         & 3.86
E+03                      & 6.82
E+03                                \\\hline
$f_{27}$                           & 2.00
E+03                      & 6.77
E+02                      & 6.95
E+02                       & 9.01
E+02                       & \textbf{5.00
E+02}             & 5.00
E+02                         & 5.26
E+02                      & 1.16
E+03                                \\\hline
$f_{28}$                           & 8.44
E+03                      & 5.17
E+02                      & 6.25
E+02                       & 5.16
E+03                       & \textbf{2.11
E+02}             & \textbf{2.28
E+02}                & 4.62
E+02                      & 3.28
E+03                                \\\hline
$f_{29}$                           & 9.34
E+03                      & \textbf{9.27
E+02}             & 1.25
E+03                       & 3.21
E+03                       & \textbf{1.02
E+03}             & 1.21
E+03                         & 1.18
E+03                      & 2.56
E+03                                \\\hline
$f_{30}$                           & 1.89
E+09                      & 8.91
E+05                      & \textbf{8.63
E+05}              & 1.30
E+08                 & 7.02
E+06                      & 1.68
E+06                         & \textbf{6.59
E+05}             & 7.87
E+07  \\
\hline\hline
\end{tabularx}
\vspace{5pt}
\end{table}
\FloatBarrier
\begin{table}[h]
\caption{Average Rankings achieved by Friedman test for CEC 2017 benchmark functions for $D=50$. The bold font indicates the smaller value.\label{tab2CEC50}}
\newcolumntype{C}{>{\centering\arraybackslash}X}
\begin{tabularx}{\textwidth}{|C|C|C|C|}
\hline\hline
\textbf{Algorithm}	& \textbf{Average Ranking}	& \textbf{Normalized} & \textbf{Ranks}\\\hline
ABC           & 8.00             & 3.23       & 8          \\\hline
PSO           & 2.97          & 1.20        & 2          \\\hline
TLBO          & 3.17          & 1.28       & 3          \\\hline
Jaya          & 6.86          & 2.77       & 7          \\\hline
GWO           & 3.90           & 1.57       & 5          \\\hline
GWO-DE        & \textbf{2.48} & \textbf{1.00} & \textbf{1} \\\hline
jDE           & 3.31          & 1.33       & 4          \\\hline
\emph{DE/best/1/bin} & 5.31          & 2.14       & 6 \\
\hline\hline
\end{tabularx}
\vspace{5pt}
\end{table}
\FloatBarrier
Figure \ref{fig:boxCEC50} shows the box plots for this case. We notice that GWO-DE performs better in the execept of $f_{11}$ where jDE has the smaller distribution. However, in this case GWO-DE performs closely to jDE better than the remaining algorithms.  PSO and TLBO perform well in most cases.

\begin{figure}[H]%
\centering
\subfigure[\label{fig:firstA2}]{\includegraphics[height=1.5in]{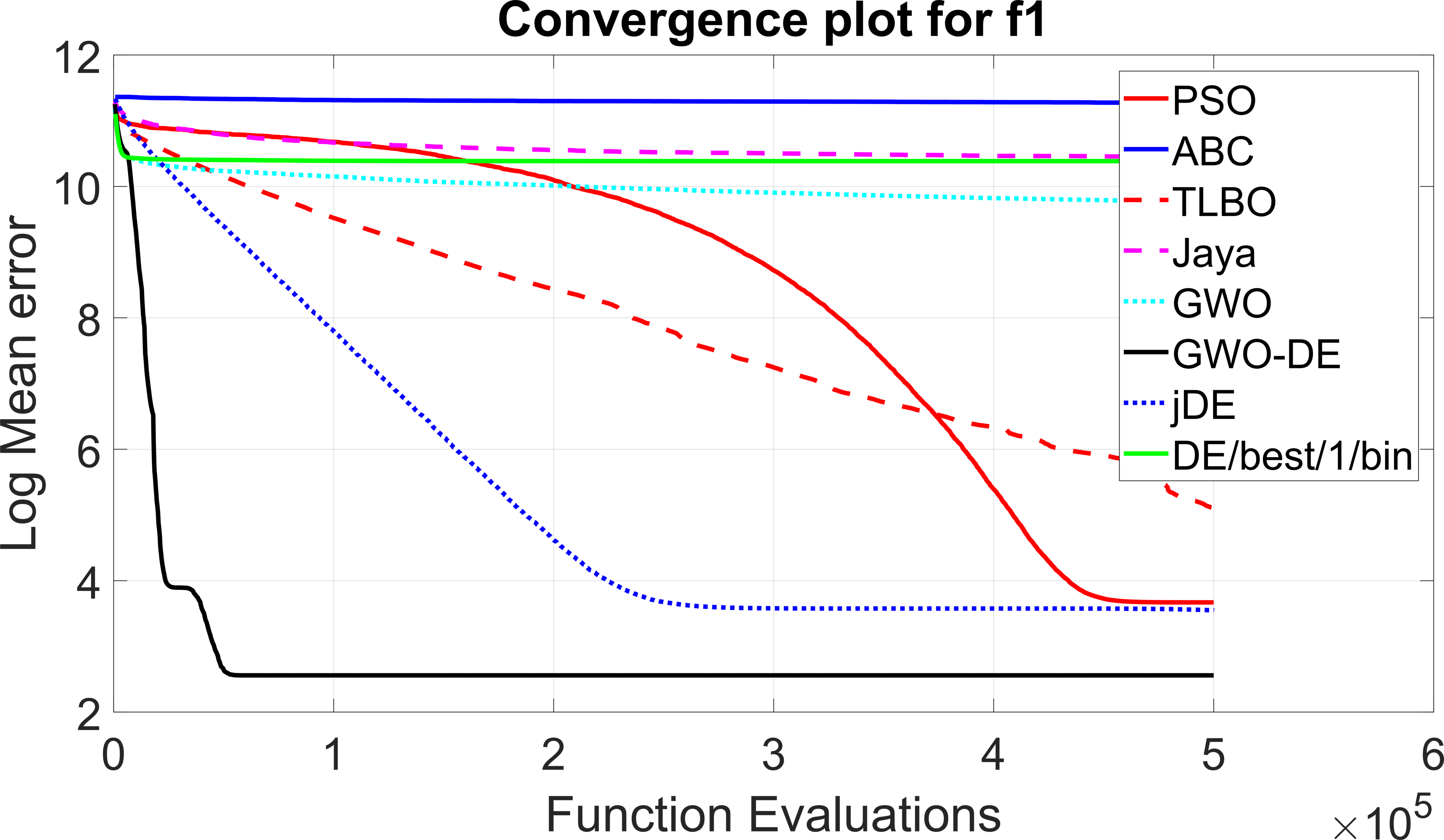}}%
\hfill
\subfigure[\label{fig:firstB2}]{\includegraphics[height=1.5in]{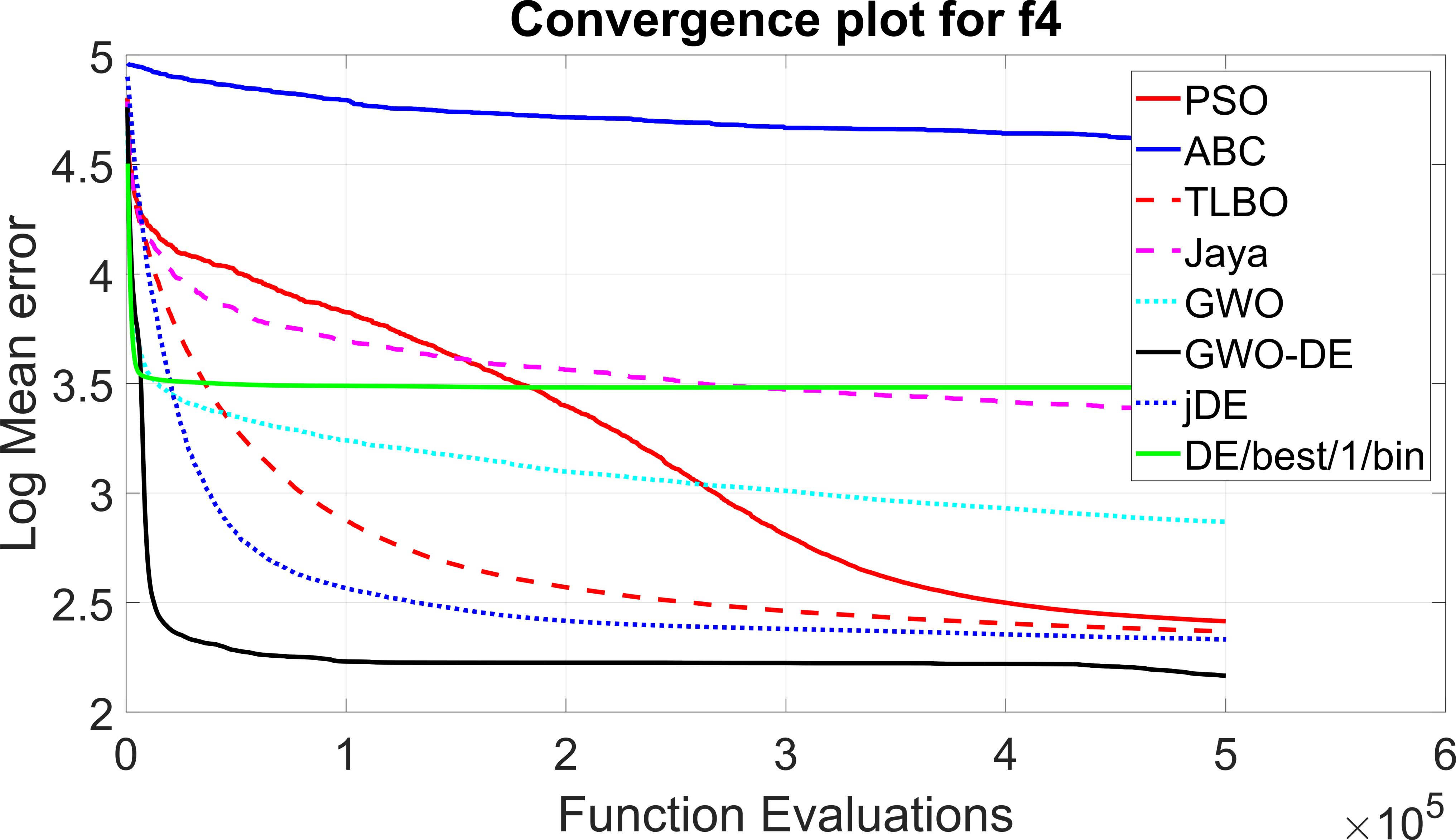}}%
\hfill
\subfigure[\label{fig:firstC2}]{\includegraphics[height=1.5in]{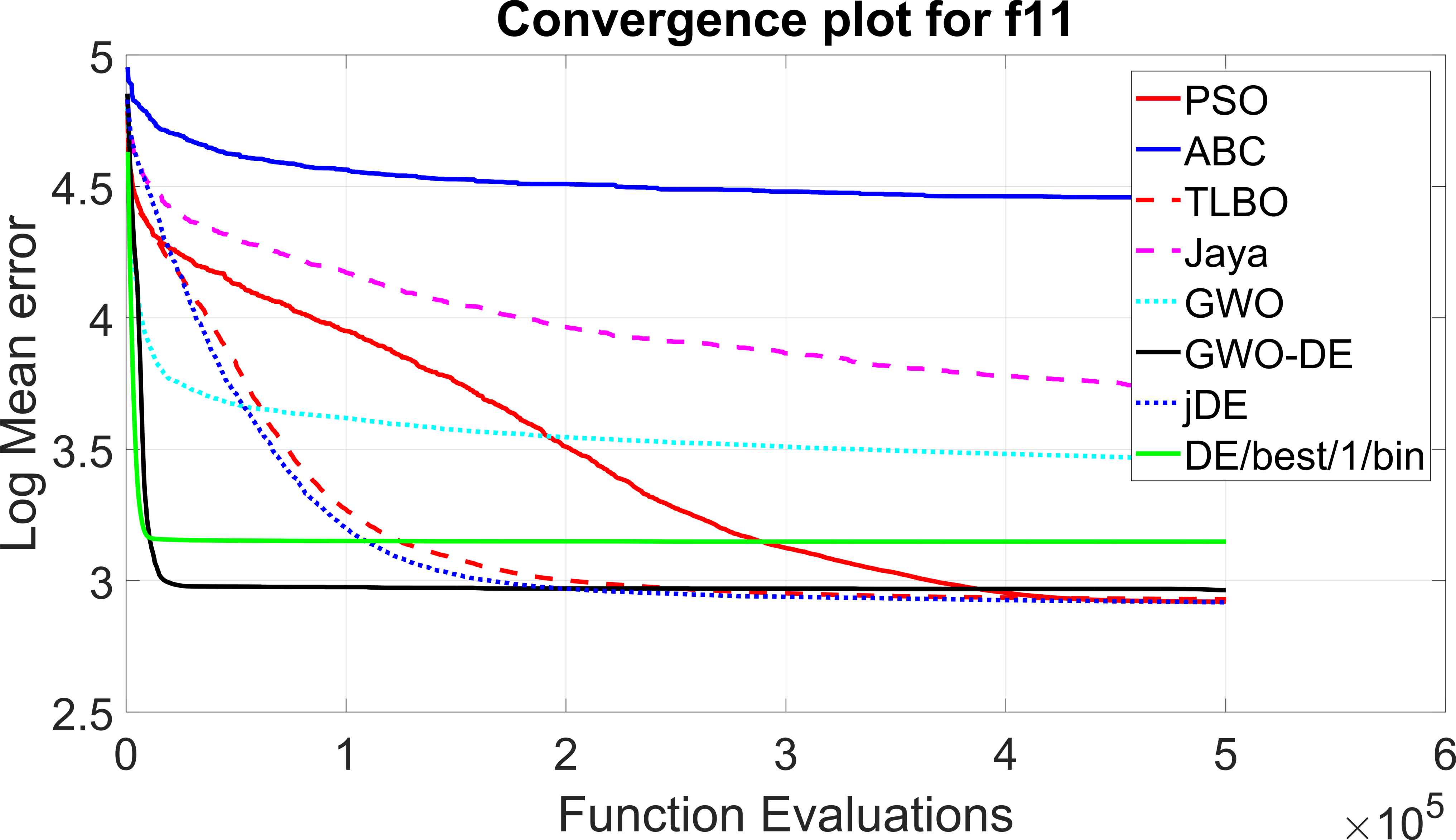}}%
\hfill
\subfigure[\label{fig:firstD2}]{\includegraphics[height=1.5in]{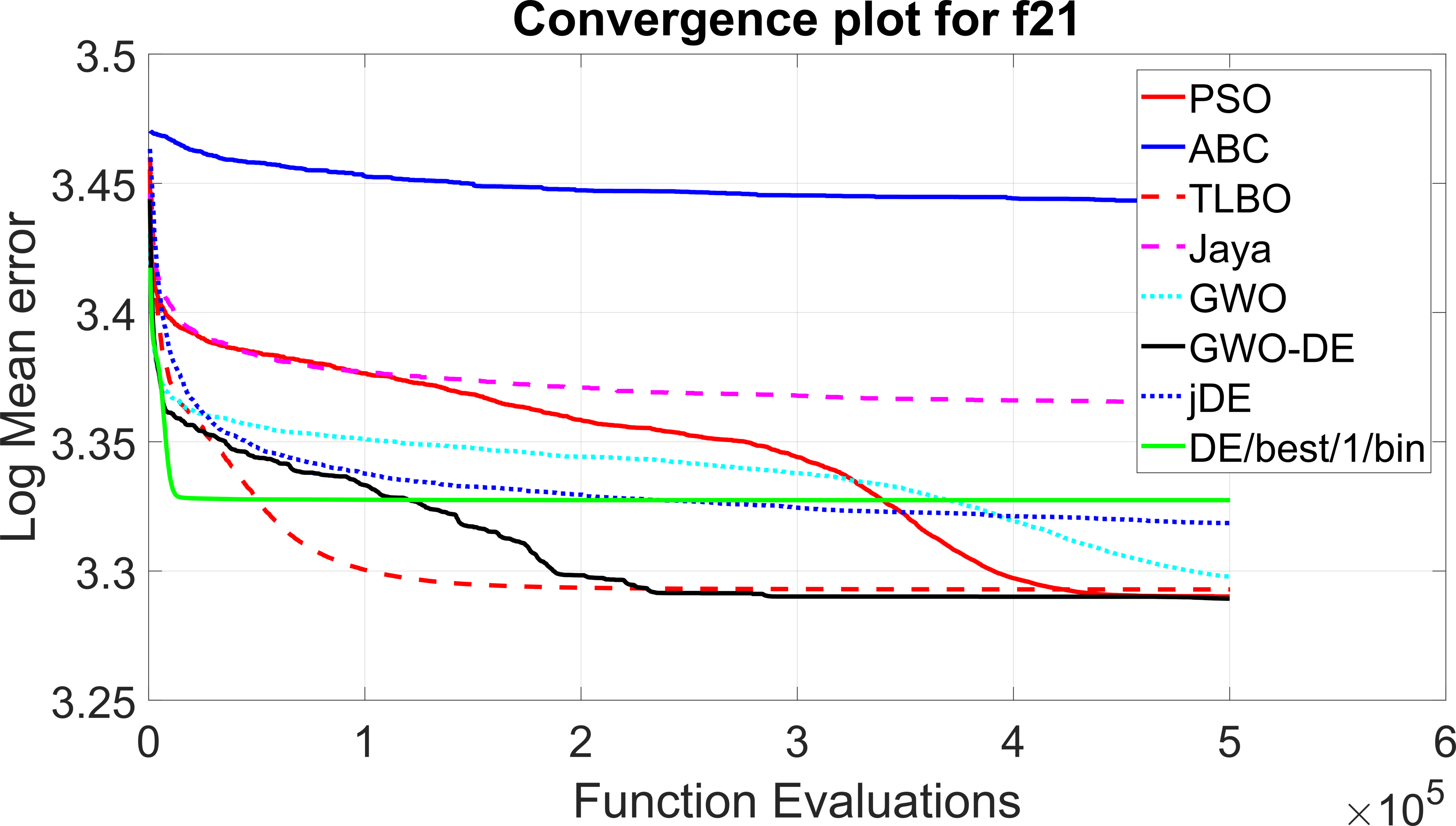}}%
\caption{Convergence rate plots of log mean error distribution for CEC 2017 benchmark functions, $D=50$, a) $f_1$, b) $f_4$, c) $f_{11}$, d) $f_{21}$.}
\label{fig:convCEC50}%
\end{figure}

Figure \ref{fig:convCEC50} shows the convergence diagrams for the same functions used in the previous subsection. GWO-DE is also very satisfactory in terms of convergence rate. We notice that GWO-DE converges faster to the final solution compared to competition in the majority of reference functions. Only in $f_{21}$ case, \emph{DE/best/1/bin} converges faster than GWO-DE however, the final value is larger than GWO-DE.

\FloatBarrier
\FloatBarrier
\newpage

\subsubsection{100 Dimensions}

\begin{figure}[H]%
\centering
\subfigure[\label{fig:firstA4}]{\includegraphics[height=1.5in]{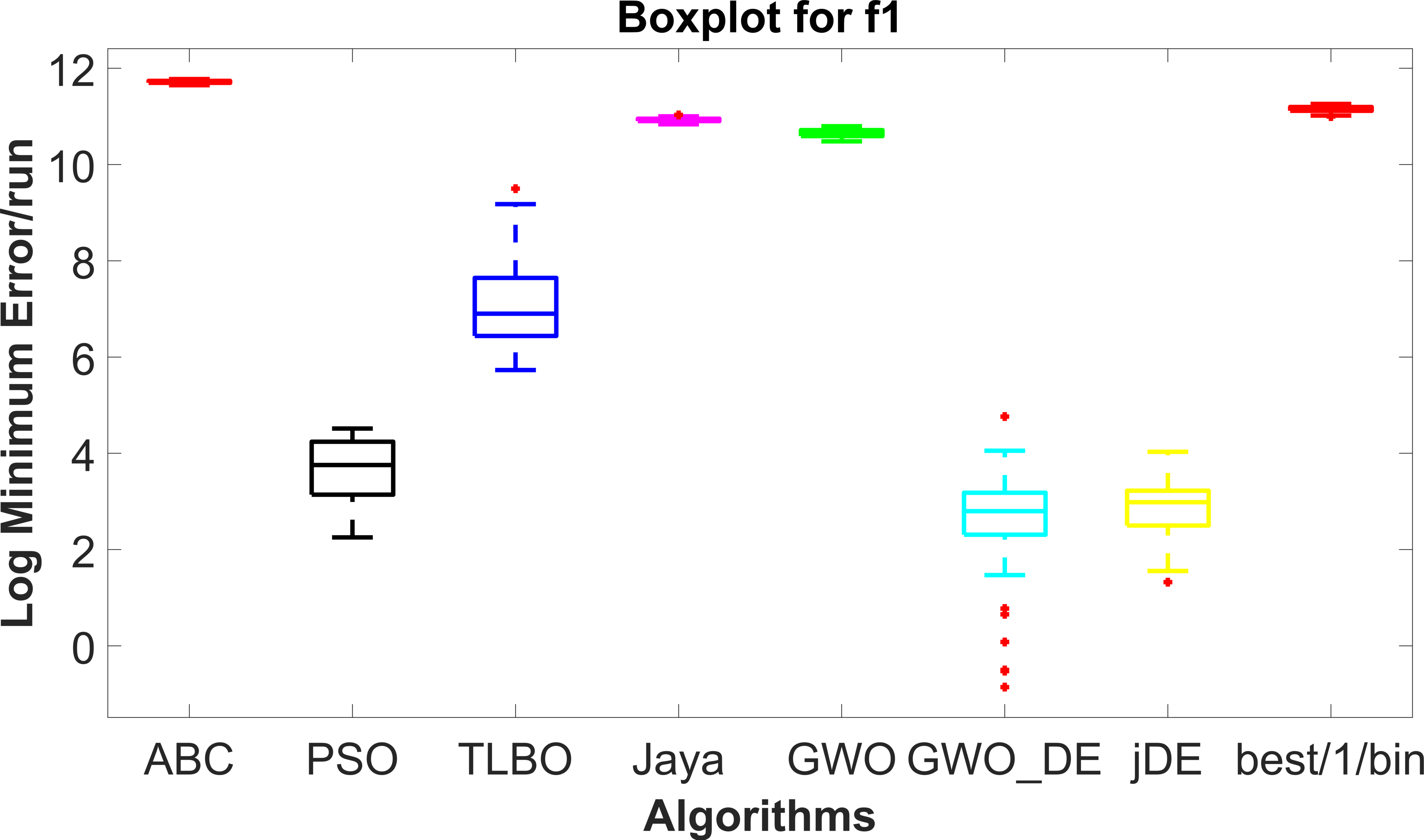}}%
\hfill
\subfigure[\label{fig:firstB4}]{\includegraphics[height=1.5in]{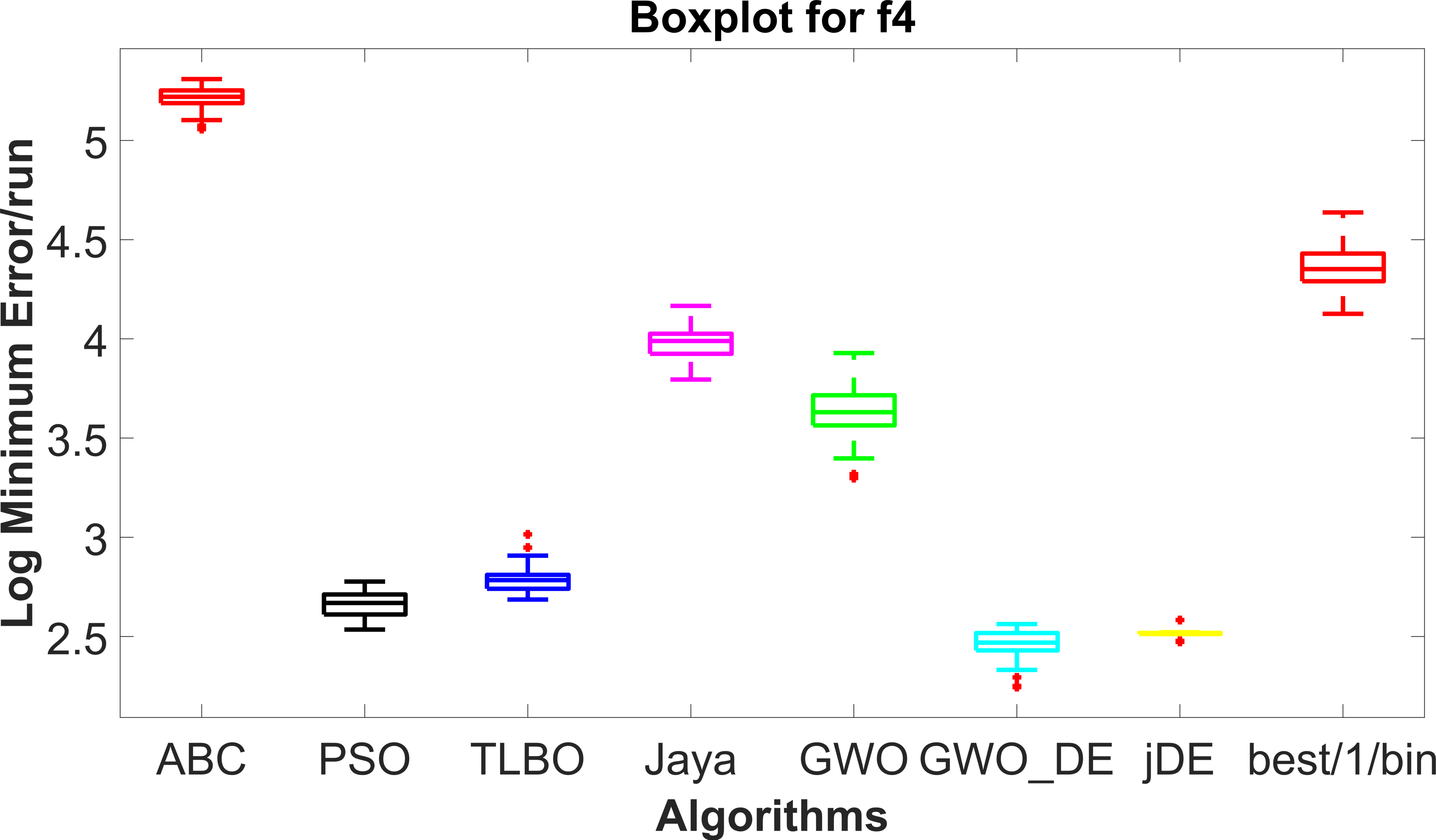}}%
\hfill
\subfigure[\label{fig:firstC4}]{\includegraphics[height=1.5in]{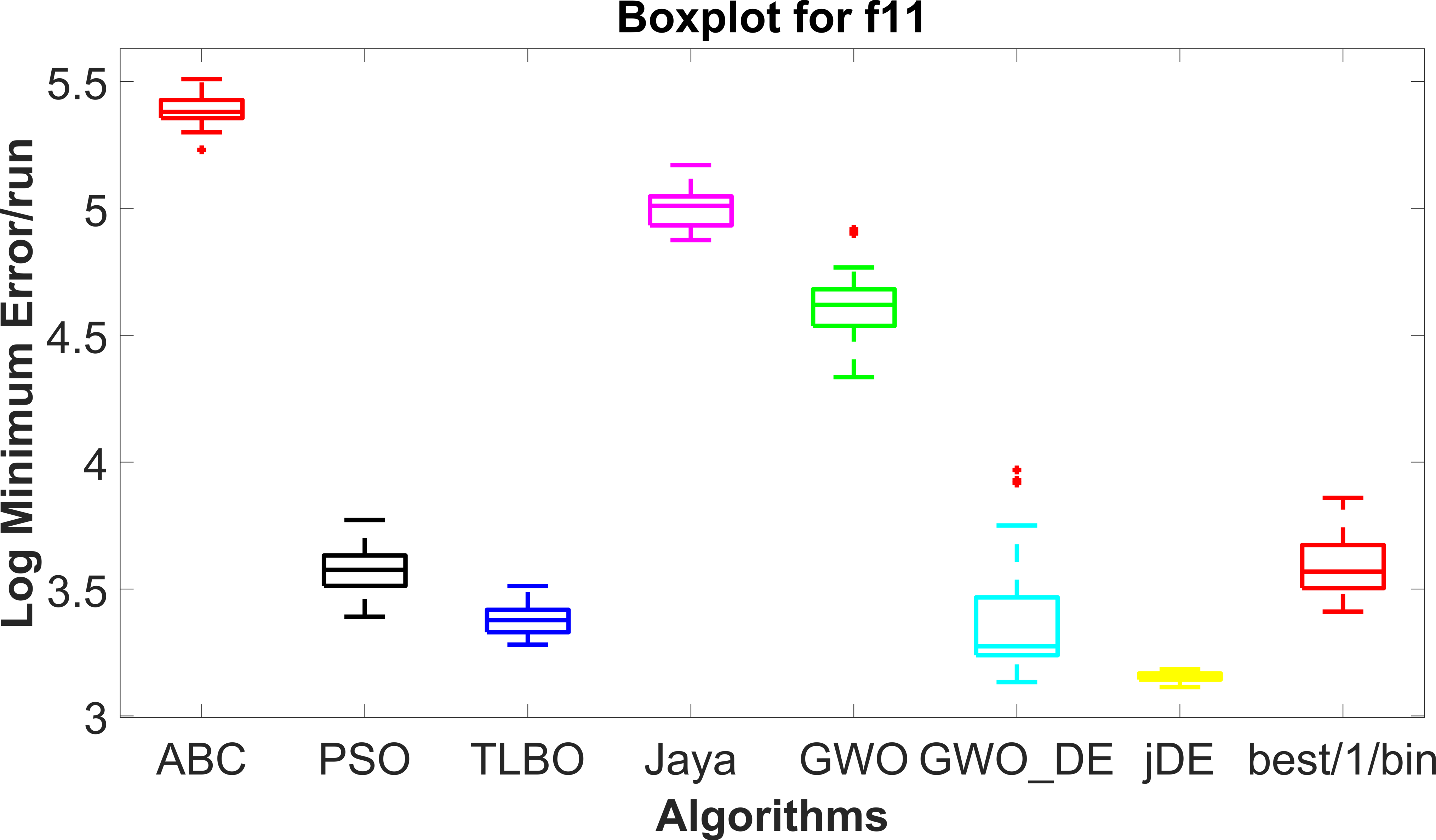}}%
\hfill
\subfigure[\label{fig:firstD4}]{\includegraphics[height=1.5in]{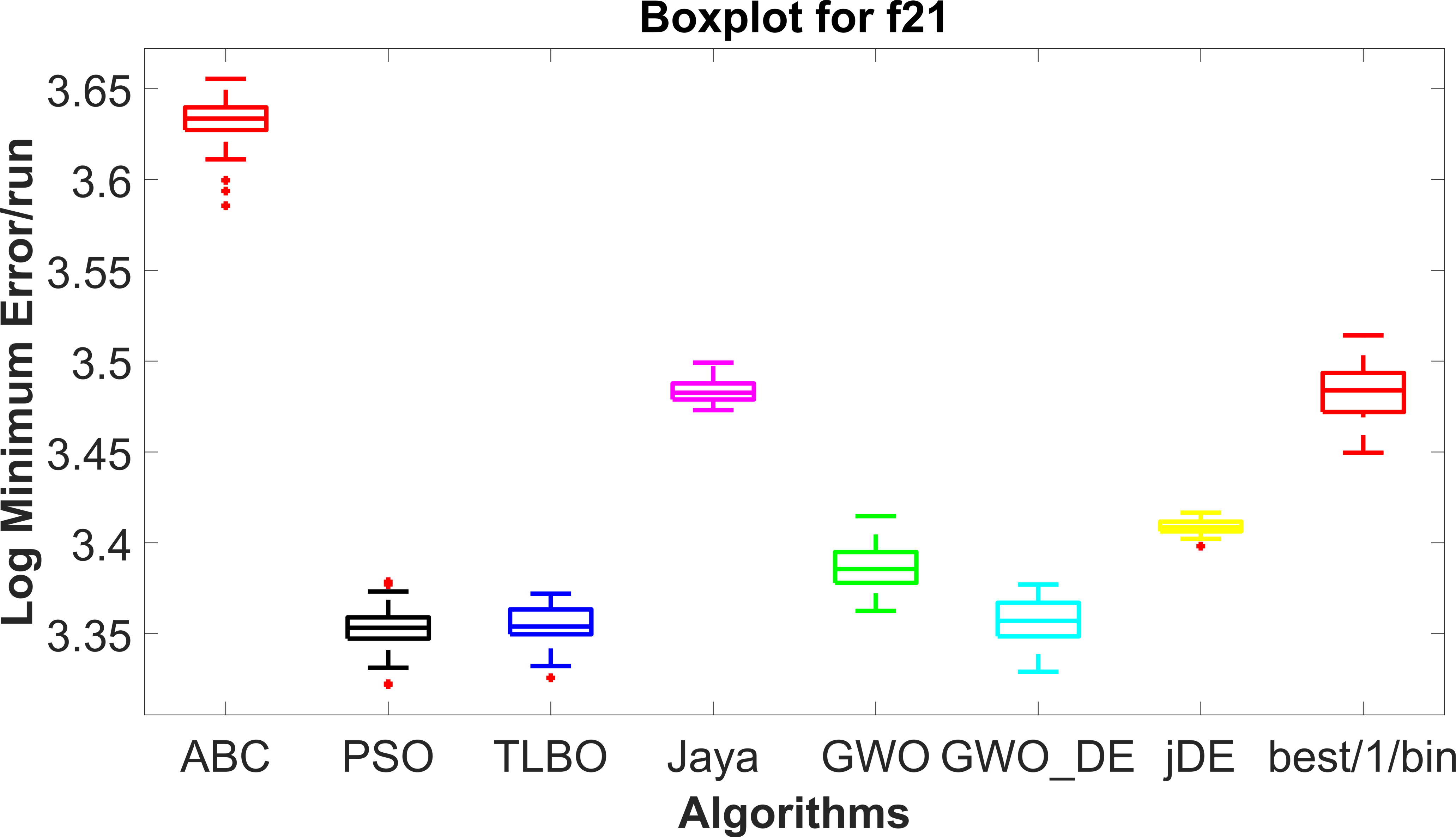}}%
\caption{Box plots of log error distribution for CEC 2017 benchmark functions, $D=100$, a) $f_1$, b) $f_4$, c) $f_{11}$, d) $f_{21}$.}
\label{fig:boxCEC100}%
\end{figure}

\begin{figure}[H]%
\centering
\subfigure[\label{fig:firstA5}]{\includegraphics[height=1.5in]{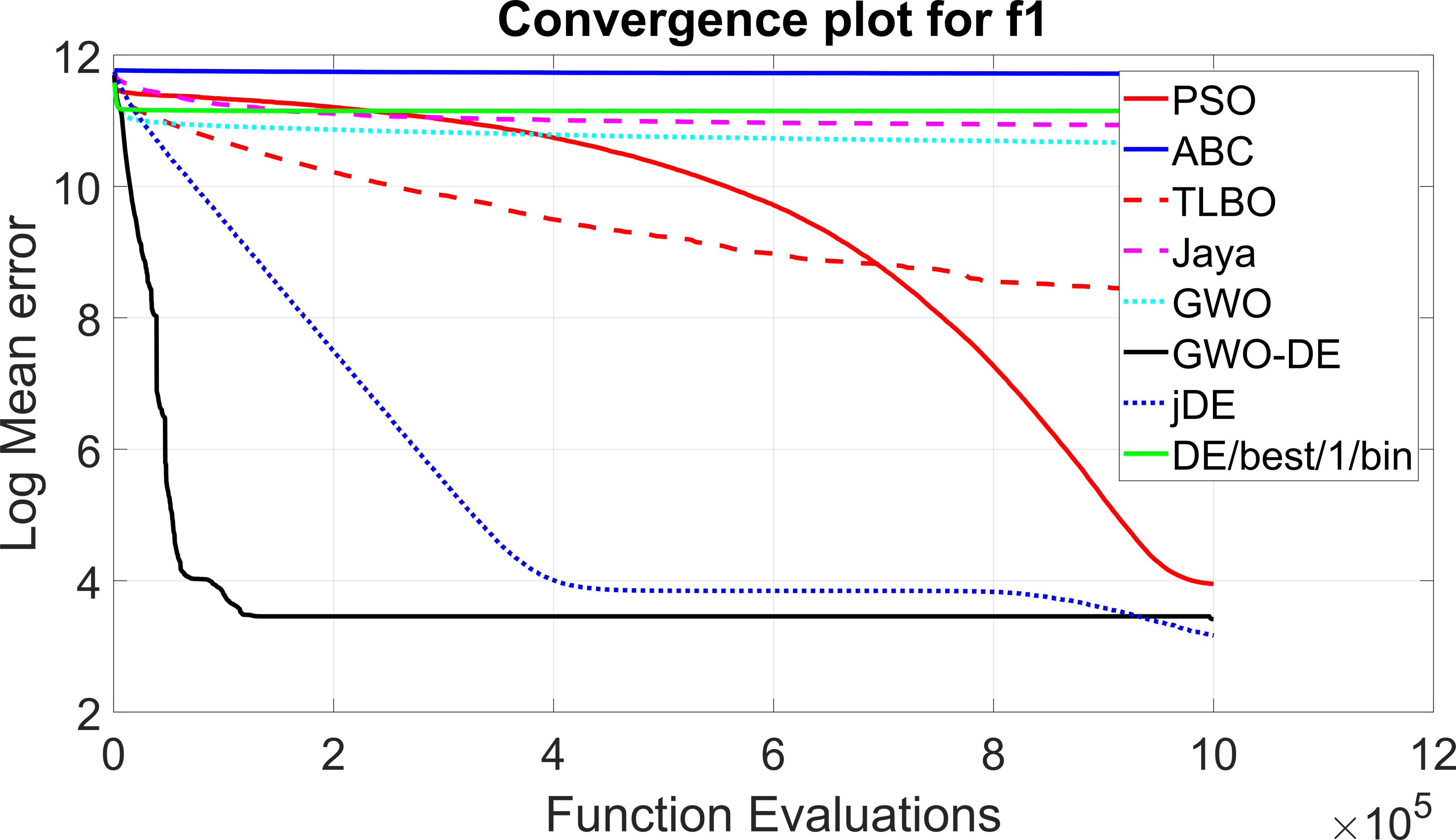}}%
\hfill
\subfigure[\label{fig:firstB5}]{\includegraphics[height=1.5in]{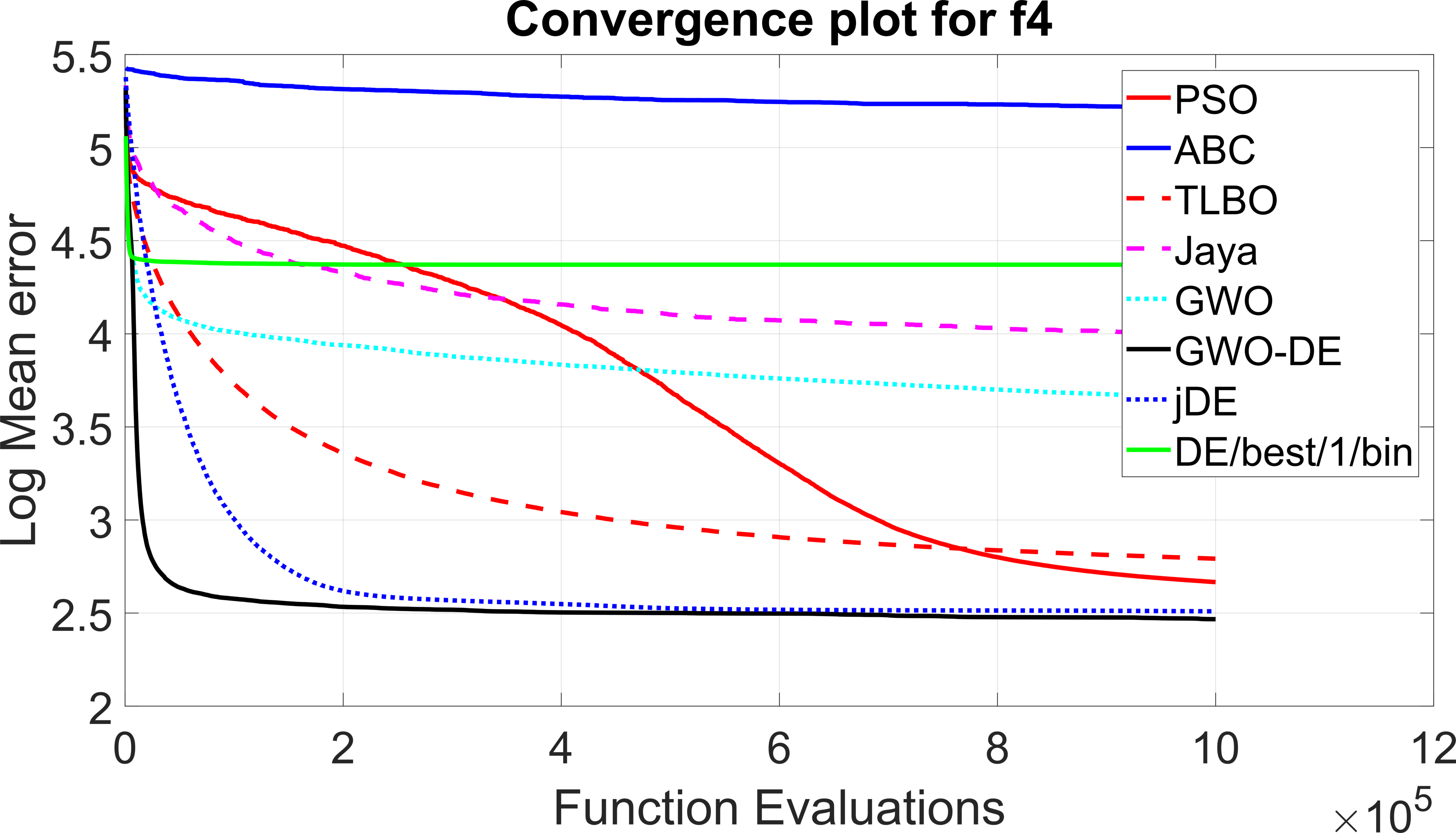}}%
\hfill
\subfigure[\label{fig:firstC5}]{\includegraphics[height=1.5in]{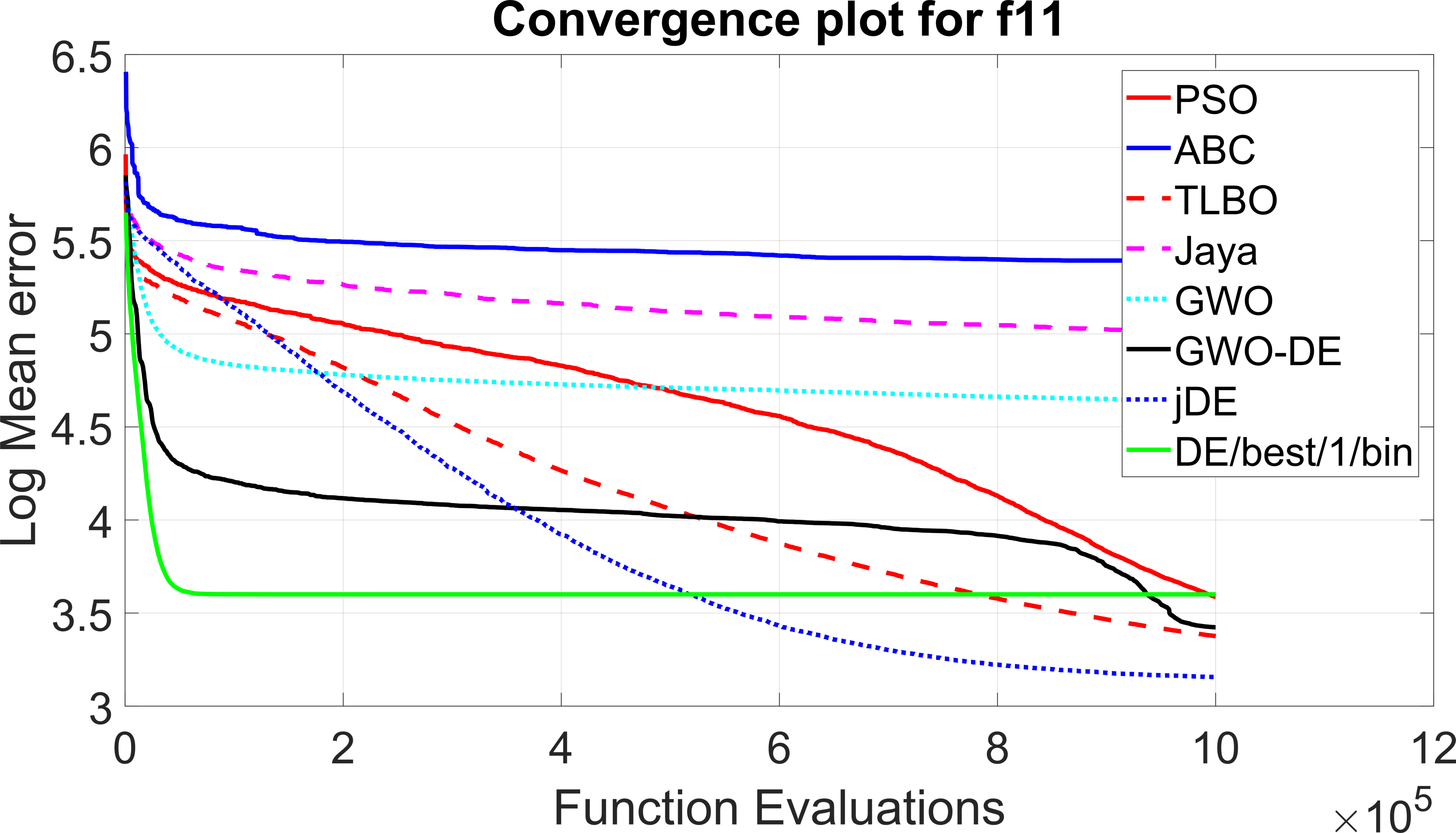}}%
\hfill
\subfigure[\label{fig:firstD5}]{\includegraphics[height=1.5in]{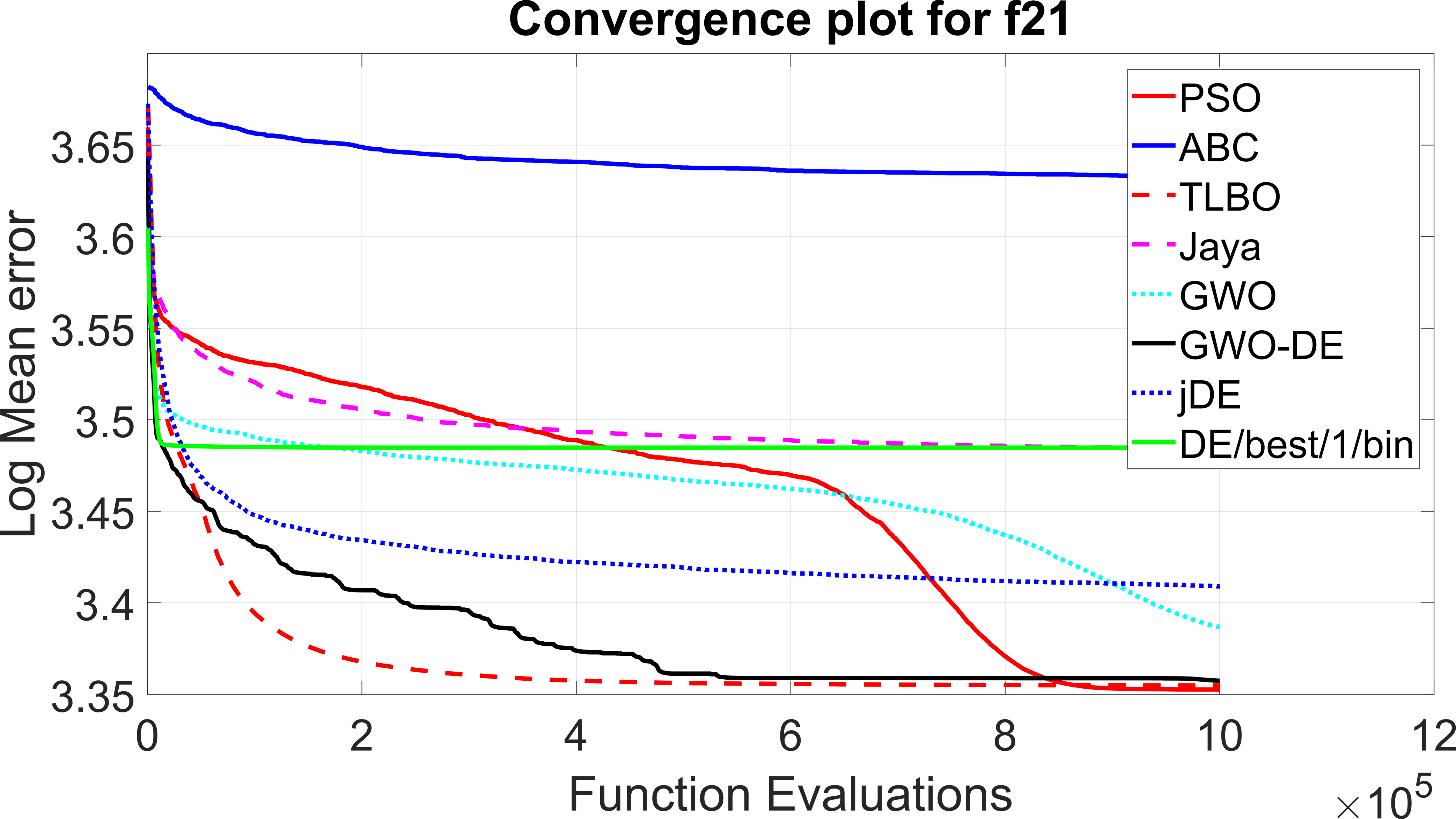}}%
\caption{Convergence rate plots of log mean error distribution for CEC 2017 benchmark functions, $D=100$, a) $f_1$, b) $f_4$, c) $f_{11}$, d) $f_{21}$.}
\label{fig:convCEC100}%
\end{figure}

Finally, we increase the CEC 2017 problem dimensions to $100$.
Again, each algorithm was run independently $50$ times per case, with a population size equal to $200$ and a maximum number of objective function evaluations equal to $1,000,000$. Therefore, the difficulty of the problem has increased significantly.
The comparative results are listed in Table \ref{tab1CEC100} and Table \ref{tab2CEC100}.
Table \ref{tab1CEC100} shows that GWO-DE has the smallest average error value in $8$ of the $28$ problems, which matches the jDE performance. However, this is more than any other algorithm. On the other hand, jDE shows the best or second-best performance in $16$ problems, while GWO-DE in $14$. However, it can be considered that GWO-DE is again the overall best optimizer, since it ranks first of the Friedman test as reported in Table \ref{tab2CEC100}.
The distribution of the mean error values is illustrated in Figure \ref{fig:boxCEC100}. We notice that GWO-DE has the smaller error values in most cases, while jDE is better in one case.
Figure \ref{fig:convCEC100} shows the convergence plots. GWO-DE maintains satisfactory performance in terms of convergence speed. In this case \emph{DE/best/1/bin} performs fast in terms of convergence speed, however, it does not obtain the smallest error value.
\begin{table}[h]
\caption{Comparative results of CEC 2017 benchmark functions for $D=100$. The bold font indicates the smaller value.\label{tab1CEC100}}
\newcolumntype{C}{>{\centering\arraybackslash}X}
\begin{tabularx}{\textwidth}{|C|C|C|C|C|C|C|C|C|}
\hline\hline
\textbf{Test   function}     & \textbf{ABC} & \textbf{PSO}      & \textbf{TLBO} & \textbf{Jaya} & \textbf{GWO}      & \textbf{GWO-DE}   & \textbf{jDE}      & \textbf{{DE/
best/
1/
bin}} \\
\hline
$f_1$                              & 5.18
E+11                      & 8.94
E+03                      & 1.63
E+08                       & 8.55
E+10                       & 4.53
E+10                      & \textbf{2.58
E+03}                & \textbf{1.47
E+03}             & 1.41
E+11                                \\\hline
$f_3$                              & 5.80
E+05                      & 2.36
E+05                      & 3.06
E+05                       & 6.05
E+05                       & 2.15
E+05                      & 1.44
E+05                         & 6.29
E+05                      & \textbf{2.81
E+04}                       \\\hline
$f_4$                              & 1.65
E+05                      & 3.64
E+02                      & 5.19
E+02                       & 9.57
E+03                       & 4.35
E+03                      & \textbf{1.93
E+02}                & 2.23
E+02                      & 2.34
E+04                                \\\hline
$f_5$                              & 2.34
E+03                      & \textbf{4.49
E+02}             & 5.96
E+02                       & 1.24
E+03                       & 6.15
E+02                      & \textbf{4.32
E+02}                & 7.33
E+02                      & 9.84
E+02                                \\\hline
$f_6$                              & 1.48
E+02                      & 1.40
E+01                      & 2.36
E+01                       & 6.91
E+01                       & 3.39
E+01                      & 3.15
E+01                         & \textbf{6.70
E-06}             & 6.24
E+01                                \\\hline
$f_7$                              & 1.13
E+04                      & 8.83
E+02                      & 9.39
E+02                       & 2.18
E+03                       & 1.03
E+03                      & \textbf{8.84
E+02}                & \textbf{8.39
E+02}             & 2.97
E+03                                \\\hline
$f_8$                              & 2.48
E+03                      & \textbf{4.30
E+02}             & 6.32
E+02                       & 1.30
E+03                       & 6.46
E+02                      & \textbf{4.27
E+02}                & 7.25
E+02                      & 1.06
E+03                                \\\hline
$f_9$                              & 1.41
E+05                      & 1.56
E+04                      & 9.94
E+03                       & 5.14
E+04                       & 2.73
E+04                      & 1.10
E+04                         & \textbf{0.00
E+00}             & 3.74
E+04                                \\\hline
$f_{10}$                           & 3.05
E+04                      & 1.80
E+04                      & 2.04
E+04                       & 3.02
E+04                       & 1.70
E+04                      & \textbf{1.32
E+04}                & 2.49
E+04                      & 1.83
E+04                                \\\hline
$f_{11}$                           & 2.45
E+05                      & \textbf{3.16
E+03}             & 1.68
E+03                       & 1.00
E+05                       & 4.24
E+04                      & 1.95
E+03                         & \textbf{7.32
E+02}             & 3.29
E+03                                \\\hline
$f_{12}$                           & 2.66
E+11                      & 1.64
E+07                      & \textbf{1.41
E+07}              & 2.42
E+10                       & 8.83
E+09                      & \textbf{3.24
E+06}                & 4.09
E+06                      & 3.05
E+10                                \\\hline
$f_{13}$                           & 6.94
E+10                      & 1.07
E+04                      & \textbf{6.24
E+03}              & 2.89
E+09                       & 5.59
E+08                      & 1.36
E+04                         & \textbf{3.12
E+03}             & 2.04
E+09                                \\\hline
$f_{14}$                           & 1.61
E+08                      & 1.06
E+06                      & 4.40
E+05                       & 2.36
E+07                       & 4.42
E+06                      & 9.89
E+04                         & \textbf{1.83
E+04}             & 3.34
E+04                                \\\hline
$f_{15}$                           & 3.26
E+10                      & 7.07
E+03                      & \textbf{9.35
E+02}              & 1.24
E+09                       & 1.06
E+08                      & 3.40
E+03                         & \textbf{3.56
E+03}             & 9.94
E+07                                \\\hline
$f_{16}$                           & 2.34
E+04                      & 3.60
E+03                      & \textbf{3.97
E+03}              & 9.83
E+03                       & 4.05
E+03                      & \textbf{3.54
E+03}                & 6.15
E+03                      & 5.73
E+03                                \\\hline
$f_{17}$                           & 5.54
E+06                      & \textbf{2.76
E+03}             & \textbf{2.78
E+03}              & 9.32
E+03                       & \textbf{2.91
E+03}             & 3.12
E+03                         & 3.82
E+03                      & 4.31
E+03                                \\\hline
$f_{18}$                           & 1.81
E+08                      & 4.07
E+06                      & 5.51
E+05                       & 4.36
E+07                       & 5.06
E+06                      & 4.82
E+05                         & \textbf{2.61
E+05}             & \textbf{7.02
E+04}                       \\\hline
$f_{19}$                           & 2.93
E+10                      & 1.00
E+04                      & \textbf{1.20
E+03}              & 1.39
E+09                       & 1.09
E+08                      & 5.74
E+03                         & \textbf{3.41
E+03}             & 1.26
E+08                                \\\hline

\end{tabularx}
\vspace{5pt}
\end{table}
\FloatBarrier
\begin{table}[h]
\newcolumntype{C}{>{\centering\arraybackslash}X}
\begin{tabularx}{\textwidth}{|C|C|C|C|C|C|C|C|C|}
\hline

$f_{20}$                           & 6.20
E+03                      & \textbf{3.08
E+03}             & \textbf{1.94
E+03}              & 5.20
E+03                       & 2.90
E+03                      & 2.35
E+03                         & 3.71
E+03                      & 2.99
E+03                                \\\hline
$f_{21}$                           & 2.68
E+03                      & \textbf{6.52
E+02}             & 6.64
E+02                       & 1.45
E+03                       & 8.37
E+02                      & \textbf{6.77
E+02}                & 9.64
E+02                      & 1.45
E+03                                \\\hline
$f_{22}$                           & 2.98
E+04                      & 2.04
E+04                      & \textbf{7.39
E+03}              & 3.08
E+04                       & 1.89
E+04                      & 1.48
E+04                         & 2.56
E+04                      & 1.94
E+04                                \\\hline
$f_{23}$                           & 3.13
E+03                      & \textbf{8.89
E+02}             & \textbf{1.02
E+03}              & 1.86
E+03                       & 1.28
E+03                      & 1.11
E+03                         & 1.12
E+03                      & 2.18
E+03                                \\\hline
$f_{24}$                           & 4.85
E+03                      & \textbf{1.38
E+03}             & 1.64
E+03                       & 2.40
E+03                       & 1.88
E+03                      & \textbf{1.83
E+03}                & 1.55
E+03                      & 3.56
E+03                                \\\hline
$f_{25}$                           & 3.55
E+04                      & 8.95
E+02                      & 1.09
E+03                       & 6.71
E+03                       & 3.04
E+03                      & \textbf{7.23
E+02}                & \textbf{7.74
E+02}             & 1.20
E+04                                \\\hline
$f_{26}$                           & 3.62
E+04                      & \textbf{8.66
E+03}             & 2.08
E+04                       & 2.03
E+04                       & \textbf{1.29
E+04}             & 1.09
E+04                         & 1.05
E+04                      & 2.64
E+04                                \\\hline
$f_{27}$                           & 3.40
E+03                      & 8.89
E+02                      & 1.16
E+03                       & 1.77
E+03                       & \textbf{5.00
E+02}             & 5.00
E+02                         & 6.28
E+02                      & 2.44
E+03                                \\\hline
$f_{28}$                           & 2.40
E+04                      & 6.96
E+02                      & 9.63
E+02                       & 1.78
E+04                       & \textbf{7.86
E+02}             & \textbf{2.52
E+02}                & 5.71
E+02                      & 1.76
E+04                                \\\hline
$f_{29}$                           & 2.06
E+05                      & \textbf{3.32
E+03}             & 3.83
E+03                       & 1.04
E+04                       & \textbf{3.10
E+03}             & 3.52
E+03                         & 4.04
E+03                      & 8.54
E+03                                \\\hline
$f_{30}$                           & 1.29
E+10                      & 1.26
E+04                      & \textbf{1.16
E+04}              & 2.12
E+09                       & 3.73
E+08                      & 9.41
E+04                         & \textbf{6.02
E+03}             & 1.13
E+09      \\
\hline\hline
\end{tabularx}
\vspace{5pt}
\end{table}

\begin{table}[h]
\caption{Average Rankings achieved by Friedman test for CEC 2017 benchmark functions for $D=100$.The bold font indicates the smaller value.\label{tab2CEC100}}
\newcolumntype{C}{>{\centering\arraybackslash}X}
\begin{tabularx}{\textwidth}{|C|C|C|C|}
\hline\hline
\textbf{Algorithm}	& \textbf{Average Ranking}	& \textbf{Normalized} & \textbf{Ranks}\\\hline
ABC           & 7.90           & 3.28       & 8          \\\hline
PSO           & 2.97          & 1.23       & 2          \\\hline
TLBO          & 3.10           & 1.29       & 3          \\\hline
Jaya          & 6.66          & 2.76       & 7          \\\hline
GWO           & 4.28          & 1.78       & 5          \\\hline
GWO-DE        & \textbf{2.41} & \textbf{1.00} & \textbf{1} \\\hline
jDE           & 3.14          & 1.30        & 4          \\\hline
\emph{DE/best/1/bin} & 5.55          & 2.30        & 6 \\
\hline\hline
\end{tabularx}
\vspace{5pt}
\end{table}
\FloatBarrier
\FloatBarrier
\FloatBarrier
\section{Conclusions}\label{gwoConcl}
In this paper, we have proposed a new hybrid GWO-DE algorithm.
In the comparative study carried out, various optimization algorithms were applied to a fairly wide range of test functions. The comparison between the algorithms was done through non-parametric statistical tests. It has been shown that the new hybrid GWO-DE algorithm that we propose is very competitive, as it is comparatively better than a multitude of other hybrid algorithms examined, as well as some algorithms already established in the literature.  The numerical results indicate that the GWO-DE algorithm outperforms the original GWO and DE algorithms.  This result is obtained for the specific reference functions used. GWO-DE appears to combine the ability to adapt and find satisfactory solutions to a wide range of multidimensional problems with a satisfactory speed of convergence.
\section*{Author Contributions}{Conceptualization, I.D.B.  and M.S.P.; methodology, I.D.B.,M.S.P., S.S., G.E.K. and A.D.B; software, I.D.B.,M.S.P., S.S., G.E.K. and A.D.B; validation, P. D., I.D.B.,M.S.P.,and A.D.B; formal analysis, P.D., I.D.B.,M.S.P.,A.D.B; investigation,S.S., G.E.K., G.K.,P.S, Z.D.Z, M.A.M.; resources,G.K.,P.S; data curation, S.K.G.,Z.D.Z; writing---original draft preparation, I.D.B.,Z.D.Z; writing---review and editing, S.K.G; visualization, I.D.B.,M.S.P. S.S., G.E.K. and A.D.B; supervision, P. S., G. K., M.A.M., and S.K.G ; project administration, S.K.G.; All authors have read and agreed to the published version of the manuscript.}
\section*{Funding}{This research received no external funding}
\section*{Data Availability Statement}{Data sharing not applicable}
\section*{Conflicts of interest}{The authors declare no conflict of interest.}
\bibliographystyle{unsrt}
\bibliography{GWODE}
\end{document}